\documentclass[journal, twoside]{IEEEtran}
\usepackage{amsmath,amssymb,amsfonts}
\usepackage[ruled,vlined, noend]{algorithm2e}
\usepackage{array}
\usepackage[caption=false,font=normalsize,labelfont=sf,textfont=sf]{subfig}
\usepackage{textcomp}
\usepackage{stfloats}
\usepackage{url}
\usepackage{verbatim}
\usepackage{graphicx}
\usepackage{cite}
\usepackage[noend]{algpseudocode}
\usepackage{float}
\usepackage{comment}
\usepackage{bm}
\usepackage{microtype}
\usepackage{physics}
\usepackage{balance}
\usepackage[colorlinks]{hyperref}
\usepackage{booktabs, multirow}

\usepackage[usernames,dvipsnames]{xcolor}

\DeclareMathOperator*{\argmin}{arg\,min}

\def\BibTeX{{\rm B\kern-.05em{\sc i\kern-.025em b}\kern-.08em
    T\kern-.1667em\lower.7ex\hbox{E}\kern-.125emX}}
\begin{document}

\title{Learning Neural Force Manifolds for \\ Sim2Real Robotic Symmetrical Paper Folding}
\author{Andrew Choi$^{*,1}$, Dezhong Tong$^{*,2}$, Demetri Terzopoulos$^{1}$, Jungseock Joo$^3$, and Mohammad Khalid Jawed$^{\dagger,2}$
\thanks{Manuscript received 6 December 2023; accepted 1 February 2024. This article was recommended for publication by Associate Editor X. Li and Editor L. Zhang upon evaluation of the reviewers' comments. This work was supported by the National Science Foundation under Grants IIS-1925360, CAREER-2047663, and OAC-2209782.}
\thanks{The authors are with the University of California, Los Angeles (UCLA), Los Angeles, CA 90095 USA.}%
\thanks{$^{1}$Andrew Choi and Demetri Terzopoulos are with the UCLA Computer Science Department (email: {\tt \footnotesize asjchoi@cs.ucla.edu; dt@cs.ucla.edu}).}%
\thanks{$^{2}$Dezhong Tong and M. Khalid Jawed are with the UCLA Department of Mechanical \& Aerospace Engineering (email:  {\tt \footnotesize tltl960308@g.ucla.edu; \\ khalidjm@seas.ucla.edu}).}%
\thanks{$^{3} $Jungseock Joo is with the UCLA Department of Communication and also with NVIDIA Corporation, Santa Clara, CA 95051 USA (email: {\tt \footnotesize jjoo@comm.ucla.edu}).} %
\thanks{$^*$ Equal contribution.}
\thanks{$^\dagger$ Corresponding author.}
\thanks{Digital Object Identifier 10.1109/TASE.2024.3366909}
}
\maketitle

\markboth{IEEE Transactions on Automation Science and Engineering. Preprint Version. Accepted Feb. 2024}
{Choi \MakeLowercase{and} Tong \MakeLowercase{et al.}: Learning Neural Force Manifolds for Sim2Real Robotic Symmetrical Paper Folding}

\begin{abstract}
Robotic manipulation of slender objects is challenging, especially when the induced deformations are large and nonlinear. Traditionally, learning-based control approaches, such as imitation learning, have been used to address deformable material manipulation. These approaches lack generality and often suffer critical failure from a simple switch of material, geometric, and/or environmental (e.g., friction) properties. This article tackles a fundamental but difficult deformable manipulation task: forming a predefined fold in paper with only a single manipulator. A sim2real framework combining physically-accurate simulation and machine learning is used to train a deep neural network capable of predicting the external forces induced on the manipulated paper given a grasp position. We frame the problem using scaling analysis, resulting in a control framework robust against material and geometric changes. Path planning is then carried out over the generated ``neural force manifold'' to produce robot manipulation trajectories optimized to prevent sliding, with offline trajectory generation finishing 15$\times$ faster than previous physics-based folding methods. The inference speed of the trained model enables the incorporation of real-time visual feedback to achieve closed-loop model-predictive control. Real-world experiments demonstrate that our framework can greatly improve robotic manipulation performance compared to state-of-the-art folding strategies, even when manipulating paper objects of various materials and shapes.
\end{abstract}

\def\abstractname{Note to Practitioners}
\begin{abstract}
This article is motivated by the need for efficient robotic folding strategies for stiff materials such as paper.
Previous robot folding strategies have focused primarily on soft materials (e.g., cloth) possessing minimal bending resistance or relied on multiple complex manipulators and sensors, significantly increasing computational and monetary costs.   
In contrast, we formulate a robust, sim2real, physics-based method capable of folding papers of varying stiffness with a single manipulator.
The proposed folding scheme is limited to papers of homogeneous material and folding along symmetric centerlines.
Future work will involve formulating efficient methods for folding along arbitrary geometries and preexisting creases.
\end{abstract}


\section{Introduction}
\label{sec:introduction}

\begin{figure}[t!]
    \centering
	\includegraphics[width=\columnwidth]{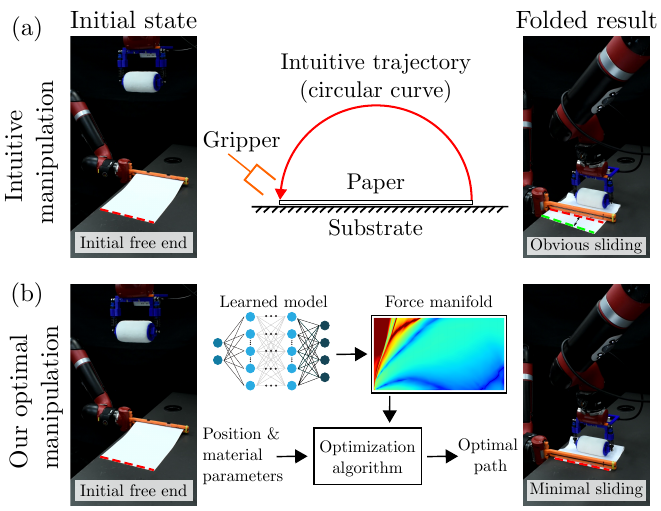}
	\caption{Half valley folding for A4 paper with (a) intuitive manipulation and (b) our designed optimal manipulation. An intuitive manipulation scheme such as tracing a semicircle experiences significant sliding due to the bending stiffness of the paper, resulting in a poor fold.
	By contrast, our optimal manipulation approach achieves an excellent fold by taking into consideration the paper's deformation to minimize sliding.}
	\label{fig:introduction}
\end{figure}

\IEEEPARstart{F}{rom} shoelaces to clothes, we encounter flexible slender structures throughout our everyday lives. These structures are often characterized by their ability to undergo large deformations when subjected even to moderate forces, such as gravity.
Therefore, the robotic manipulation of deformable objects is highly nontrivial as a robot must be able to take into account future deformations of the manipulated object in order to complete manipulation tasks successfully.

Prior research has focused primarily on manipulating either cloth or ropes~\cite{sanchez2018robotic, yin2021modeling}, and, as a result, the challenge of robotically manipulating many other deformable objects still lacks robust solutions.
This article addresses a particularly difficult deformable manipulation task --- folding paper. 
Paper is similar to cloth but typically possesses a significantly higher bending stiffness and a slippery surface. 
Therefore, when compared to folding garments and fabrics, the folding of paper requires more delicate and insightful manipulations. 
In fact, in our experiments, we observe that state-of-the-art methods for robotic fabric/cloth folding~\cite{petrik2016folding, petrik2019folding, petrik2020static} perform poorly when transferred to paper.

To tackle these challenges, we propose a framework that combines physically accurate simulation, scaling analysis, and machine learning to generate folding trajectories optimized to prevent sliding.
With scaling analysis, we make the problem non-dimensional, resulting in both dimensionality reduction and generality.
This allows us to train a single nondimensionalized neural network, whose outputs are referred to as a neural force manifold (NFM), to continuously approximate a scaled force manifold sampled purely from simulation.
Compared to numerical models that require the entire geometric configuration of the paper, NFMs map the external forces of the paper given only the grasp position.
Therefore, we can generate trajectories optimized to minimize forces (and thus minimize sliding) by applying path-planning algorithms.
Furthermore, the nondimensionality of the NFM allows us to generate trajectories for papers of various materials and geometric properties even if such parameters were not present in the training dataset. 
We show that our approach is capable of folding paper on extremely slick surfaces with little-to-no sliding (Fig.~\ref{fig:introduction}(b)).

Overall, our main contributions are as follows: 
\begin{enumerate}
    \item We formulate a solution for folding materially homogeneous sheets of paper along symmetrical centerlines in a physically robust manner using scaling analysis, resulting in complete generality concerning the modulus and density of the material, size of the paper, and environmental properties (e.g., friction).
    \item Next, we generate accurate non-dimensional simulation data to train a ``neural force manifold'' for optimal trajectory generation. We exploit the high inference speed of our trained model with a perception system to construct a robust and efficient closed-loop model-predictive control algorithm for the folding task in near real-time.
    \item Finally, we demonstrate full sim2real realization through an extensive robotic case study featuring 360+ folding experiments involving paper sheets of various materials and shapes. We compare our method against both natural paper folding strategies as well as the previous state of the art in robotic rectangular fabric folding~\cite{petrik2020static, petrik2016folding}. 
\end{enumerate}
Moreover, we offer demonstration videos and release all our code as open-source software.\footnote{See \url{https://github.com/StructuresComp/deep-robotic-paper-folding}.}

The remainder of the article is organized as follows: 
We begin with a review of related work in Sec.~\ref{sec:related-work}.
A brief description of the folding problem is presented in Sec.~\ref{sec:problem_statement}.
The formulation of a reduced-order physics-based model is discussed in Sec.~\ref{sec:physics-based-model}, where we formulate the folding problem using scaling analysis.
In Sec.~\ref{sec:deep_learning_and_optimization}, we formulate our learning framework as well as algorithms for optimal path planning.
Next, in Sec.~\ref{sec:robotic_system}, we introduce our robotic system and formulate our closed-loop visual feedback pipeline.
Experimental results for a robot case study and analysis of the results are given in Sec.~\ref{sec:experiments_and_analysis}.
Subsequently, Sec.~\ref{sec:additional_discussion} provides additional discussion regarding performing multiple folds and the importance of single-manipulator folding.
Finally, we make concluding remarks and discuss the potential of future research avenues in Sec.~\ref{sec::conclusion}.

\section{Related Work}
\label{sec:related-work}

The majority of prior work addressing the folding problem can be roughly divided into four categories: mechanical design-based solutions, vision-based solutions, learning-based solutions, and model-based solutions.

Mechanical design-based approaches typically involve tackling the folding problem using highly specialized manipulators or end effectors.
Early approaches involved specialized punches and dies for sheet metal bending \cite{kim1998automated}.
More recently, highly specialized manipulators for robotic origami folding have also been developed \cite{balkcom2008robotic}. Such methods can reliably produce repeatable folding but are often limited to a highly specific fold, geometry, and/or material.

Vision-based approaches involve folding deformable materials by generating folding motions purely from visual input. 
These techniques are commonly applied to tasks such as folding clothes, where the primary focus is on detecting garment shape or key grasp points. 
Techniques for key feature extraction involve random decision trees~\cite{doumanoglou2016folding}, 
RGB-D sensing data analysis~\cite{maitin2010cloth, twardon2015interaction},
and fitting strategies where the detected state of deformed clothes are compared against precomputed shapes~\cite{kita2011clothes, doumanoglou2014autonomous}.
Given the soft nature of clothes, subsequent manipulations are often formulated intuitively. While some prior research employs models to predict optimal manipulation sequences, these models are typically oversimplified and lack physical details~\cite{miller2012geometricfolding}.
Such approaches can be effective and rather simple to implement, but do not transfer well to paper folding as paper has a much higher stiffness than fabric and will attempt to restore its natural, undeformed state if not properly handled. 

Learning-based approaches involve the robot learning how to fold through training data.
The most popular has been to learn control policies from human demonstrations, also known as learning from demonstrations (LfD). Prior research has demonstrated flattening and folding towels \cite{lee2015learning, lee2015lfd}.
Teleop demonstrations are a popular avenue for training policies and have been used to learn how to manipulate deformable linear objects (DLOs) \cite{rambow2012autonomous} as well as folding fabric \cite{yang2017folding}.
To eliminate the need for expensive human-labeled data, researchers have also focused on tackling the sim2real problem for robotic folding, with reinforcement learning being used to train robots to fold fabrics and clothes completely from simulation \cite{petrik2019folding, matas2018sim, lin2020softgym}.
More recently, Zheng et al.~\cite{zheng2022autonomous} used reinforcement learning to train a robot to flip pages in a binder through tactile feedback.
Pure learning-based methods have shown promising performance, but only for specific tasks whose state distribution matches the training data. Such methods tend to generalize quite poorly; e.g., when the material or geometric properties change drastically. 

Model-based approaches, where the model can either be known or learned, often use model-predictive control to manipulate the deformable object. 
Learned models involve learning the natural dynamics of deformable objects through random perturbations \cite{yan2020learning}.
These models are generally fast, but they can be inaccurate when experiencing new states.
Theoretical models are often formulated to be as physically accurate as possible, which enables the direct application of their predictive power in the real world. 
Examples of this have been published for both strip folding \cite{petrik2016folding, petrik2020static} and garment folding \cite{yinxiao2015garmentfolding}. 
Physical models are often constructed using energy-based formulations~\cite{sanchez2018robotic, yin2021modeling, zhu2022challenges}, where various elastic energies are computed based on the topological properties of the simulated objects to solve their deformed shape under manipulation. 
For example, Wakamatsu and Hirai~\cite{wakamatsu2004static} modeled deformable linear objects (rods) with flexure (bending), torsion, and extension (stretching), while Jia et al.~\cite{jia2014grasping} introduced manipulation as a potential energy to compute the deformations of deformable planar objects. 
However, theoretical models are usually quite expensive to run and must often face a trade-off between accuracy and efficiency.

Despite the large quantity of prior research focusing on 2D deformable object manipulation, the majority of these efforts have limited their scope to soft materials such as towels and cloth. 
Such materials are highly compliant and often do not exhibit complicated nonlinear deformations, thus allowing for solutions lacking physical insight.
We instead tackle the scenario of folding paper of various stiffnesses with a single manipulator.
Because of its relatively high bending stiffness and slippery surface, paper is significantly more difficult to manipulate since large deformations will cause sliding of the paper on the substrate. 
Such an example can be observed in Fig.~\ref{fig:introduction}(a), where intuitive folding trajectories that may work on towels and cloth fail for paper due to undesired sliding. 

However, a few researchers have attempted to solve the paper folding problem.
For example, Elbrechter et al.~\cite{elbrechter2012folding} demonstrated paper folding using visual tracking and real-time physics-based modeling, with impressive results, but they required expensive end effectors (two Shadow Dexterous Hands), one end effector to hold the paper down while folding at all times, and the paper to have AR tags for visual tracking. Similarly, Namiki et al.~\cite{namiki2015robotic} also achieved paper folding through dynamic motion primitives and used physics-based simulations to estimate the deformation of the paper sheet, also requiring highly specialized manipulators and an end effector to hold the paper down while folding.
By contrast, our method can fold papers reliably without any need for holding down the paper during the folding operation and requires only a simple 3D printed gripper.

Other researchers have also attempted to fold with a single manipulator while minimizing sliding \cite{petrik2016folding, petrik2019folding, petrik2020static}, but their methods focused on fabrics whose ends were taped down to the substrate.
Though these methods have achieved favorable folding accuracy using a physical model for garments and fabric, we have observed in our experiments that their generated trajectories perform poorly when applied to paper folding.
We believe that this is due to their local optimization strategy of solving the subsequent grasp pose using only the current grasp.
In contrast, we generate our folding trajectories through global optimization, thus showcasing the importance of considering both current and future deformation states during the paper manipulation process.

\section{Problem Statement}
\label{sec:problem_statement}

\begin{figure}
\centerline{\includegraphics[width =\columnwidth]{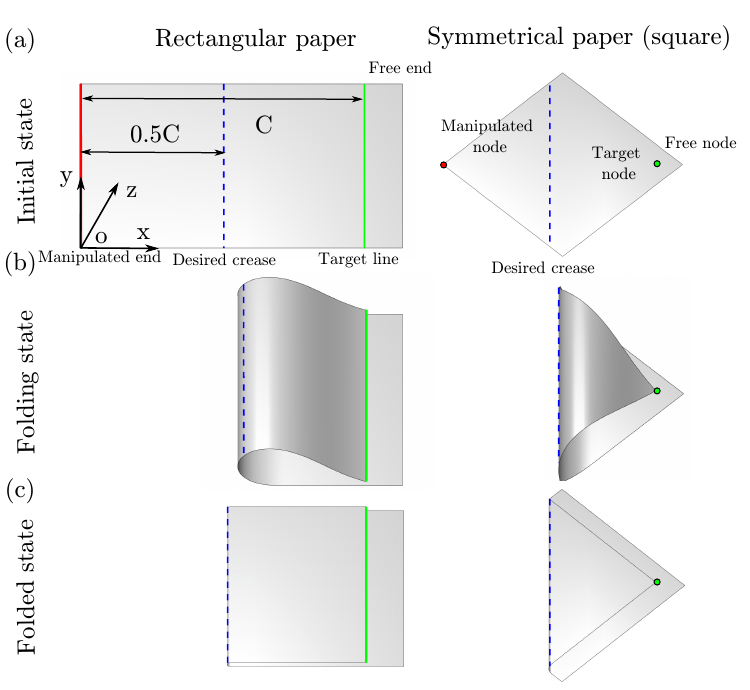}}
\caption{States of the paper during the folding process. The manipulation process involves two steps. The first (folding) step transitions the paper from the initial state (a), where the paper lies flat on the substrate, to the folding state (b), where the manipulated end is moved to the ``crease target'' line $C$. The second (creasing) step then transitions the paper from state (b) to the final folded state (c), which involves forming the desired crease on the paper.}
\label{fig::Problem_Statement}
\end{figure}

This article studies a simple yet challenging task in robotic folding: creating a predefined crease on a sheet of paper of typical symmetrical geometry (e.g., rectangular, diamond, etc.) as illustrated in Fig.~\ref{fig::Problem_Statement}.
Only one end of the paper is manipulated while the other end is left free. Thus, extra fixtures are unnecessary and the folding task can be completed by a single manipulator, which simplifies the workspace, but slippage of the paper against the substrate must be mitigated during manipulation, which presents a challenge.

The task can be divided into two steps.
The first is manipulating one end of the paper from the initial flat state (Fig.~\ref{fig::Problem_Statement}(a)) to the folding state (Fig.~\ref{fig::Problem_Statement}(b)), with the goal that the manipulated edge or point should overlap precisely with the crease target line or point $C$ as shown in the figure. 
In the second step, the paper is then permanently deformed to form the desired crease at $C/2$, thus achieving the final folded state (Fig.~\ref{fig::Problem_Statement}(c)). 

As creasing the paper is trivial, the main challenge lies in minimizing the displacement of the free end of the paper during the first step. 
The paper's large nonlinear deformations and slippery surface make accurate predictions of the folding paper's status crucial for minimizing displacement. Since permanent deformations are absent in the first step, we model the paper's nonlinear deformations using a 2D planar rod model with a linear elastic assumption, which is discussed in detail in the next section. 
This physical model is then combined with scaling analysis and machine learning to generate physically-informed folding trajectories optimized to minimize sliding.
With the first step concluded, simple motion primitives are used to complete the final paper creasing.

\section{Physics-based Model and Analysis}
\label{sec:physics-based-model}

We next present the numerical framework for studying the underlying physics of the paper folding process. 
First, we analyze the main deformations of the manipulated paper and prove that a 2D model is sufficient to learn the behaviors of the manipulated paper so long as the sheet is symmetrical. 
Second, we briefly introduce a physically accurate numerical model based on prior work in the field of computer graphics~\cite{bergou2008der}. 
Third, we formulate a generalized strategy for paper folding using scaling analysis.

\subsection{Reduced-Order Model Representation}
\label{sec:reduced_order_model}

Paper is a unique deformable object. Unlike cloth, its surface is developable~\cite{hilbert2021geometry}; i.e., the surface can bend but not stretch. 
Furthermore, shear deformations are not of particular importance as paper possesses a negligible thickness-to-length ratio.
Therefore, the primary nonlinear deformation when folding paper in our scenario is bending deformation.
We postulate that the nonlinear behaviors of paper arise primarily from a balance of bending and gravitational energies: $\epsilon_b \sim \epsilon_g$.

To further understand the energy balance of the manipulated paper, we analyze a finite element of the paper, as shown in Fig.~\ref{fig::DER}(b). 
The bending energy of this piece can be written as
\begin{equation}
 \epsilon_b = \frac{1}{2} k_b \kappa^2 l,
 \label{eq::bending}
 \end{equation}
where $l$ is its undeformed length of the piece, $\kappa$ is its curvature, and its bending stiffness is
 \begin{equation}
    k_b =  \frac{1}{12} E w h^3,
\end{equation}
where $w$ is its undeformed width, $h$ is its thickness, and $E$ is its Young's modulus. 
The gravitational potential energy of the piece is
\begin{equation}
    \epsilon_g = \rho w h l g H,
\label{eq::gravitational}
\end{equation}
where $\rho$ is its volume density, and $H$ is its vertical height above the rigid substrate.

From the above equations, we obtain a characteristic length called the gravito-bending length, which encapsulates the influence of bending and gravity:
\begin{equation}
L_{gb} = \left(\frac{E h^2}{24 \rho g} \right)^\frac{1}{3} \sim \left(\frac{H}{\kappa^2} \right)^\frac{1}{3}.
\label{eq::Lgb}
\end{equation}
The length is in units of meters, and we can observe that it scales proportionally to the ratio of vertical height to curvature squared, which are the key quantities describing the deformed configuration of the manipulated paper. 
Note that the formulation of $L_{gb}$ contains only one geometric parameter, the paper thickness $h$, which means that other geometric quantities (i.e., length $l$ and width $w$) have no influence on the deformed configuration.

Additionally, due to the symmetrical geometry and material homogeneity of the paper, the curvature $\kappa$ should be identical for all regions at the same height $H$. 
Therefore, we can simply use the centerline of the paper, as shown in Fig.~\ref{fig::DER}(a), to express the paper's configuration. 
We model this centerline as a 2D planar rod since deformations are limited to the $x$-$z$ plane, and 
implement a discrete differential geometry (DDG) numerical simulation to simulate the 2D planar rod. 
The next section presents the details of this numerical framework.

\subsection{Discrete Differential Geometry Numerical Model}
\label{sec::DDG_framework}

\begin{figure}
\centerline{\includegraphics[width =\columnwidth]{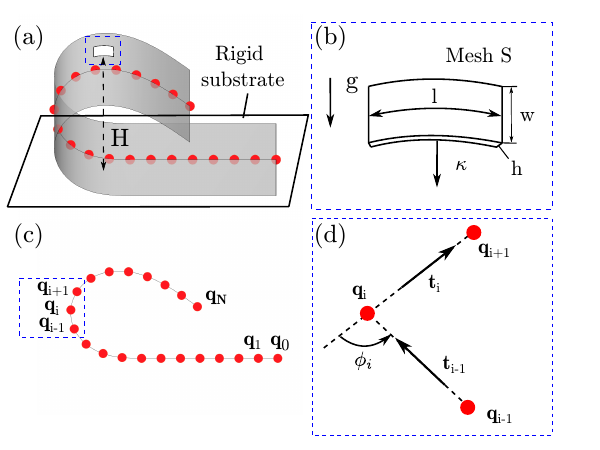}}
\caption{(a) Schematic of paper during the folding state. (b) Bending deformations of a small piece in the paper. (c) Reduced-order discrete model (planer rod) representation of the paper. (d) Notations in the discrete model.}
\label{fig::DER}
\end{figure}

Subsequent to the pioneering work on physics-based modeling and simulation of deformable curves, surfaces, and solids in computer graphics \cite{terzopoulos1987elastically,terzopoulos1988modeling,terzopoulos1988deformable}, the community has shown impressive results using DDG-based simulation frameworks.
For example, the Discrete Elastic Rods (DER)~\cite{bergou2008der} framework has shown remarkable efficiency and physically validated accuracy when simulating deformable linear objects for a wide variety of scenarios including elastic coiling~\cite{jawed2014coiling}, deployment~\cite{tong2023sim2real}, helix bifurcation~\cite{tong2021automated}, overhand knot tightening~\cite{choi2021implicit, tong2023fully}, overhand knot buckling~\cite{tong2023snapbuckling}, flagella buckling~\cite{jawed2015propulsion, tong2023fully}, and dynamic cantilever beams~\cite{choi2024dismech}.  
Given this success, we too choose to use DER to model the centerline of the paper as a 2D planar rod undergoing bending deformations. 

As shown in Fig.~\ref{fig::DER}(c), the discrete model is comprised of $N+1$ nodes, $\mathbf q_i$ ($0 \le i \le N$). Each node $\mathbf q_i$ represents two degrees of freedom (DOF): position along the $x$ and the $z$ axes. This results in a $2N+2$-sized DOF column vector $\mathbf q = [\mathbf q_0, \mathbf q_1, ..., \mathbf q_N]^T$ representing the configuration of the paper sheet.
Initially, all the nodes of the paper are located in a line along the $x$-axis in the paper's undeformed state.
As the robotic manipulator imposes boundary conditions on the end node $\mathbf q_N$, portions of the paper deform against the substrate, as shown in Fig.~\ref{fig:schematic}(a).
We compute the DOFs as a function of time $\mathbf q(t)$  by integrating the equations of motion at each DOF.

Before describing the equations of motion, we first outline the elastic energies of the rod as a function of $\mathbf q$. 
Kirchhoff's rod theory tells us that the elastic energies of a rod can be divided into stretching $E_s$, bending $E_b$, and twisting $E_t$ energies.
First, the stretching elastic energy is
\begin{equation}
    E_s = \frac{1}{2} k_s \sum_{i=0}^{N-1} \left( 1 - \frac{\| \mathbf q_{i+1} - \mathbf q_i \| }{\Delta l} \right)^2 \Delta l,
    \label{eq:stretching}
\end{equation}
where $k_s=EA$ is the stretching stiffness, $E$ is Young's modulus, $A = wh$ is the cross-sectional area, and $\Delta l$ is the undeformed length of each edge (segment between two nodes). 
The bending energy is
\begin{equation}
    E_b = \frac{1}{2} k_b \sum_{i=2}^{N-1} \left( 2 \tan \frac{\phi_i}{2} - 2 \tan \frac{\phi^0_i}{2} \right)^2 \frac{1}{\Delta l} ,
    \label{eq:bending}
\end{equation}
where $k_b = \frac{E wh^3}{12}$ is the bending stiffness, $w$ and $h$ are the width and thickness, respectively, $\phi_i$ is the ``turning angle'' at a node (Fig.~\ref{fig::DER}(d)), and $\phi^0_i$ is the undeformed turning angle ($0$ for paper). 
Finally, since deformations are limited to a 2D plane, we can ignore twisting energies.
The total elastic energy is therefore $E_{el} = E_s + E_b$. 

Indeed, a ratio $k_s/k_b \sim 1/h^2 \gg 1$ indicates that stretching strains will be minimal, which matches our intuition as paper is usually easy to bend but not stretch. 
Therefore, the stretching energy item in (\ref{eq:stretching}) acts as a constraint to prevent obvious stretching for the modeled planar rod.

We can now write the equations of motion as a simple force balance
\begin{equation} 
\mathbb{M} \ddot{\mathbf q}  + \frac{\partial E_{el}}{\partial \mathbf{q}} - \mathbf{F}^{\textrm{ext}} = 0,
\label{eq::force_balance}
\end{equation}
where $\mathbb M$ is the diagonal lumped mass matrix, the dots denote time derivatives of $\mathbf q$, $\frac{\partial E_{el}}{\partial \mathbf{q}}$ is the elastic force vector, and $\mathbf{F}^{\textrm{ext}}$ are the external forces acting on the paper. 
Newton's method can be used to solve (\ref{eq::force_balance}), allowing for full simulation of the 2D planar rod under manipulation.

\subsection{Generalized Solution and Scaling Analysis}
\label{subsec:scaling_analysis}

\begin{figure}
\centerline{\includegraphics[width =\columnwidth]{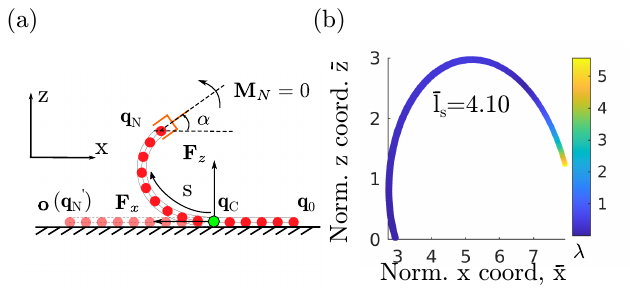}}
\caption{(a) Side view of a symmetrical paper during folding with coordinate frame and relevant notations. (b) Sampled $\lambda$ forces for a particular $\bar l_s$ of 4.10. This shows one of the sampled ``partial'' force manifolds that we use to train our neural network on.}
\label{fig:schematic}
\end{figure}

As mentioned in Sec.~\ref{sec:problem_statement}, the core of the folding task is to manipulate the end $\mathbf q_N$ to the target position $C$ starting from an initially flat state shown in Fig.~\ref{fig:schematic}(a).
To do so, we analyze the physical system in order to achieve a solution capable of minimizing sliding during manipulation.

We first denote several quantities to describe the deformed configuration of the paper.
We introduce a point $\mathbf q_C$, which is the node that connects the suspended ($z > 0$) and contact regions ($z = 0$) of the paper.
We focus solely on the suspended region as deformations occur primarily in this region.
An origin $\mathbf o$ is defined for our 2D plane which is located at the initial manipulated end $\mathbf q_N$, as shown in Fig.~\ref{fig:schematic}(a). 
For the manipulated end, the robot end-effector imposes a position $\mathbf q_N = (x, z)$ and an orientation angle $\alpha$ to control the pose of the manipulated end (Fig.~\ref{fig:schematic}(a)). We impose a constraint that the curvature at the manipulated end is always zero so that sharp bending deformations are prevented, which is crucial to preventing permanent deformations during the folding process.
On the connective node $\mathbf q_C$, the tangent is always along the $x$-axis.  
With these definitions, we can now modify (\ref{eq::force_balance}) with the following constraints:
\begin{equation}
\begin{aligned} 
\mathbb{M} \ddot{\mathbf q}  + \frac{\partial E_{el}}{\partial \mathbf{q}} - \mathbf{F}^{\textrm{ext}} &= 0,\\
 \textrm{such that} \quad  \mathbf q_N &= (x, z),\\
 \dv{\mathbf q_C}{s} &= (-1, 0), \\
 M_N & = 0,\\
 l_s & \equiv \int_{\mathbf q_C}^{\mathbf q_N} \textrm{d}s = \mathbf q_C \cdot \hat{\mathbf x},
\end{aligned}
\label{eq::Constraint}
\end{equation}
where $M_N$ is the external moment applied on the manipulated end, $s$ is the arc length of the paper's centerline, and $l_s$ is the arc length of the suspended region (from $\mathbf q_C$ to $\mathbf q_N$).

We can solve (\ref{eq::Constraint}) with the numerical framework presented in Sec.~\ref{sec::DDG_framework} resulting in a unique DOF vector $\mathbf q$. 
Note that when $\mathbf q$ is determined, we can then obtain the external forces from the substrate along the paper $\mathbf F_{\textrm{substrate}} = \mathbf F_x + \mathbf F_z$, orientation angle $\alpha$ of the manipulated end, and the suspended length $l_s$.
Recall that through (\ref{eq::Lgb}), Young's modulus $E$, thickness $h$, and density $\rho$ were determined to be the main material and geometric properties of the paper. 
Therefore, we can outline the following physical relationship relating all our quantities:
\begin{equation}
\begin{aligned}
    \lambda &= \frac{\lVert \mathbf F_x \rVert}{\lVert \mathbf F_z \rVert}, \\
    (\lambda, \alpha, l_s) &= f(E, h, \rho, x, z),
\end{aligned}
\label{eq:mapping}
\end{equation}
where $f$ is an unknown relationship. 
It is then trivial to see that to prevent sliding the relationship 
\begin{equation}
\begin{aligned}
    \lambda \leq \mu_s
\end{aligned}
\label{eq:sliding}
\end{equation}
must be satisfied, where $\mu_s$ is the static friction coefficient between the paper and the substrate.
Therefore, a trajectory that minimizes sliding is one that minimizes $\lambda$ along its path.

One glaring problem remains in that the relation $f$ must be known to generate any sort of trajectory.
In the absence of an analytical solution, the numerical framework from Sec. \ref{sec::DDG_framework} can be used to exhaustively find mappings between the inputs and outputs of $f$.
However, generating tuples in this fashion requires solving the high-dimensional problem in (\ref{eq::Constraint}). 
Such a method would be horribly inefficient and would make real-time operation infeasible.
Instead, we opt to obtain a regression approximation of $f$ by fitting a neural network on simulation data.
This approach has several shortcomings, however.
For one, directly learning $f$ is time-consuming given that (\ref{eq:mapping}) is a high-dimensional mapping that depends on five parameters as input. 
Furthermore, since the formulation directly depends on intrinsic parameters of the paper ($E$, $\rho$, and $h$), an enormously exhaustive range of simulations must be run to gather enough data to accurately learn $f$.

To proceed, we reduce the dimensionality of the problem by applying scaling analysis. 
According to the Buckingham $\pi$ theorem, we construct five unitless groups: $\bar x = x/L_{gb}$; $\bar z = z/L_{gb}$; $\bar l_s = l_s/L_{gb}$; $\alpha$, and $\lambda = F_t/F_n$, where $L_{gb}$ is the gravito-bending length (\ref{eq::Lgb}).
This results in the following unitless formulation of (\ref{eq:mapping}):
\begin{equation}
\begin{aligned}
    (\lambda, \alpha, \bar l_s) &= \mathcal{F} \left(\bar x,  \bar z \right).\\
\end{aligned}
\label{eq:mapping_without_unit}
\end{equation}
Note that the mapping $\mathcal F$ is now independent of quantities with units; e.g., material and geometric properties of the paper.
As the dimensionality of our problem has been reduced significantly, we can now express $\lambda$ as a function of just two parameters $\bar x, \bar z$.
Therefore, training a neural network to model $\mathcal F$ is now trivial as non-dimensionalized simulation data from a single type of paper can be used.
Furthermore, the low dimensionality of $\mathcal F$ allows us to easily visualize the $\lambda$ landscape along a non-dimensional 2D-plane.

We will detail the steps to model $\mathcal F$ in the next section.

\section{Deep Learning and Optimization}
\label{sec:deep_learning_and_optimization}

\begin{figure*}
	\includegraphics[width=\textwidth]{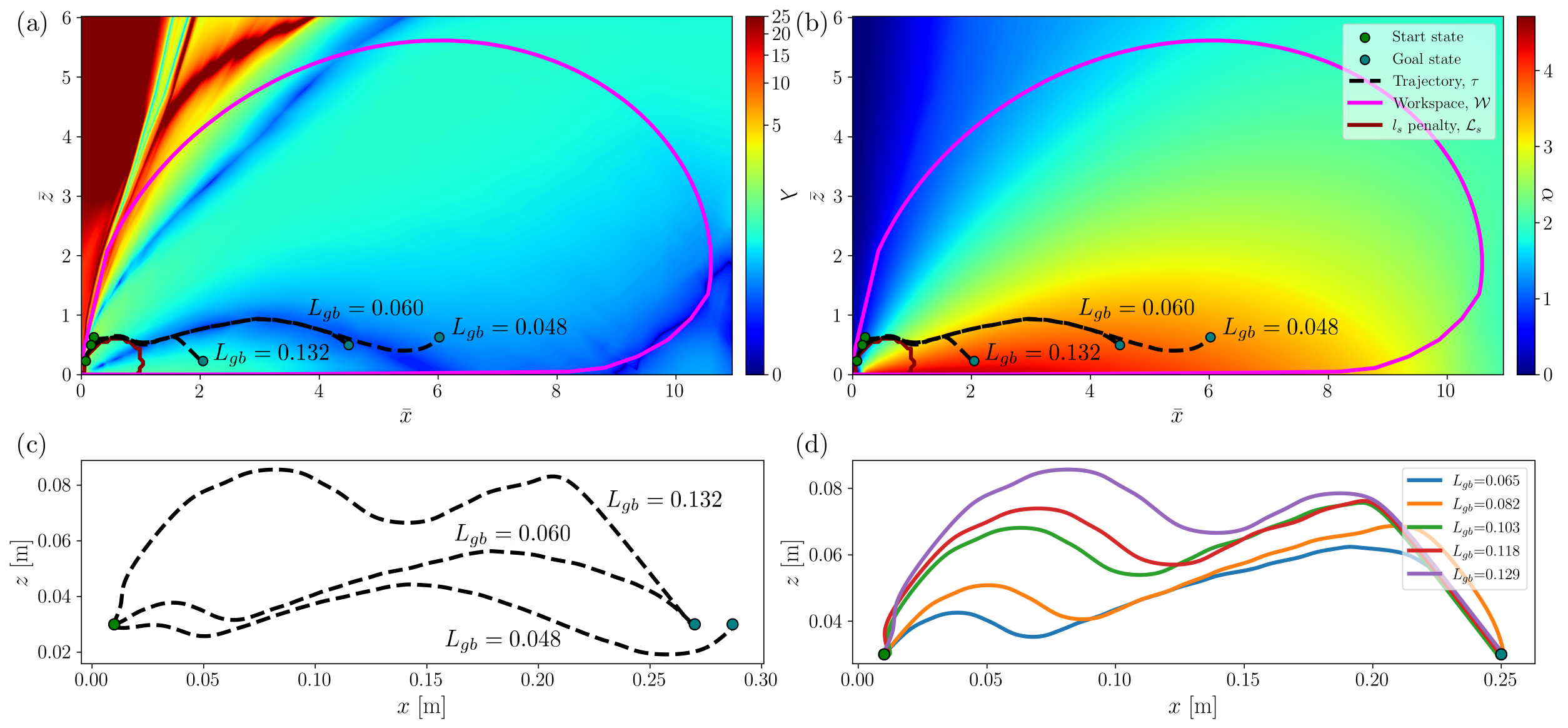}
	\caption{Visualization of the trained neural network's non-dimensionalized $\lambda$ force manifold $\mathcal M$ (a) and $\alpha$ manifold (b). An extremely low $\bar \delta$ discretization is used to showcase smoothness. 
    For the force manifold, we observe two distinctive local minima canyons.
    Note that regions outside the workspace $\mathcal W$ are physically inaccurate, but are of no consequence as they are ignored.
    For the $\alpha$ manifold, we observe continuous smooth interpolation throughout, which is crucial for producing feasible trajectories.
	Both manifolds showcase the trajectories used in the experiments for folding paper in half for $L_{gb} \in [0.048, 0.060, 0.132]$. (c) The three trajectories in (a) and (b) scaled back to real space. These are the actual trajectories used by the robot. (d) Arbitrary trajectories for various $L_{gb}$ with identical start and goal states, highlighting the effect of the material property on our control policy.}
	\label{fig:neural_network}
\end{figure*}

\subsection{Data Generation}

To learn the force manifold, we solve (\ref{eq::Constraint}) for many sampled $(x, z)$ points. An example of the partial force manifold produced from this sampling can be observed for a single suspended length in Fig.~\ref{fig:schematic}(b).
For a specific $(x, z)$ location, we apply incremental rotations along the y-axis and find the optimal rotation angle $\alpha$ that results in $M_N = 0$ on the manipulated end. 
For a particular configuration $(x, z, \alpha)$, we then record the suspended length $l_s$ as well as the tangential and normal forces experienced on the clamped end. This leads to a training dataset $\mathcal D$ consisting of six element tuples $(F_t, F_n, \alpha, l_s, x, z)$.
We then non-dimensionalize this dataset to the form $(\lambda, \alpha, \bar l_s, \bar x, \bar z)$. 

With our simulation framework, we generated a dataset $\mathcal D$ comprising a total of 95,796 training samples within a normalized suspended length of $\bar l_s \leq 6.84$ (which adequately covers the workspace of most papers), consuming 3.54 hours of compute time on an AMD Ryzen 7 3700X 8-core processor.

\subsection{Learning Force and Optimal Grasp Orientation}

To train a neural network model of $\mathcal{F}$,
\begin{equation}
\label{eq:neural_network}
    (\lambda, \alpha, \bar l_s) = \mathcal F_\textrm{NN}(\bar x, \bar z),
\end{equation}
we employed a simple fully-connected feed-forward nonlinear regression network with 4 hidden layers, each containing 392 units.
Aside from the final output layer, each layer is followed by rectified linear unit (ReLU) activation.
In addition, we preprocessed all inputs through the standardization
\begin{equation}
\label{eq:standardization}
    \mathbf x' = \frac{\mathbf x - \bar{\mathbf x}_\mathcal{D}}{\boldsymbol \sigma_\mathcal D},
\end{equation}
where $\mathbf x$ is the original input, $\bar{\mathbf x}_\mathcal D$ is the mean of the dataset $\mathcal D$, and $\boldsymbol \sigma_\mathcal D$ is the standard deviation of $\mathcal D$.

We used an initial 80-20 train-val split on the dataset $\mathcal D$ with a batch size of 128. Mean absolute error (MAE) was used as the training error. We alternated between stochastic gradient descent (SGD) and the Adam optimizer whenever training stalled. Furthermore, we gradually increased the batch size up to 4,096 and trained on the entire dataset once the MAE dropped below $0.001$. Using this scheme, we achieved an MAE of less than $0.0005$.

\subsection{Constructing the Neural Force Manifold}
\label{subsec:manifolds}

The neural force manifold (i.e., $\lambda$ outputs of $\mathcal F_\textrm{NN}$ for the workspace set) is discretized into a rectangular grid consisting of $\bar \delta \times \bar \delta$ blocks, where $\bar \delta = \delta / L_{gb}$.
For each of the blocks, we obtain and store a single $\lambda$ value using the midpoint of the block. This results in a discretized neural force manifold $\mathcal M$ represented as a $m \times n$ matrix.
For the purposes of path planning, we add two components to our manifold.
First, we do not allow exploration into any region not covered by our training dataset ($\bar l_s > 6.84)$.
We do so by defining a workspace $\mathcal W$ as all $(\bar x, \bar z)$ pairs within the convex hull of the input portion of the dataset $\mathcal D$.
Secondly, we also exclude regions within a certain $\bar l_s$ threshold. This is done as positions with small suspended lengths and large $\alpha$ angles may result in high curvatures that could cause collision with our gripper and/or plastic deformation, both of which we wish to avoid. We denote this region as the penalty region $\mathcal L_s$.
Fig.~\ref{fig:neural_network}(a) shows a visualization of $\mathcal M$ with the workspace $\mathcal W$ and penalty boundary $\mathcal L_s$ regions. The $\alpha$ values corresponding to the manifold are also shown in Fig.~\ref{fig:neural_network}(b).

\subsection{Path Planning over the Neural Force Manifold}
\label{subsec:path_planning}

Given the discretized manifold $\mathcal M$, we can now generate optimal trajectories through traditional path planning algorithms. 
Indeed, we find that there exist two local minima regions (dark blue in Fig.~\ref{fig:neural_network}(a)) in the neural force manifold $\mathcal M$. 
However, note that these two minima regions are not connected, which means that improper local optimization may result in undesired traversal through high force regions later.
As mentioned previously, prior mechanics-based efforts on folding shell-like structures (cloth) have used either physical simulations or energy-based optimization to compute the optimal subsequent grasp based solely on the current status of the manipulated object~\cite{petrik2016folding, petrik2020static}. 
We show that this local optimization approach performs poorly for paper folding in Sec.~\ref{sec:experiments_and_analysis}.
By contrast, we generate globally optimized trajectories that take into account both current and future states of the paper. 
To do so, we define an optimal trajectory $\tau^*$ as one that reaches the goal state while minimizing the sum of $\lambda$:
\begin{equation}
\tau^* = \argmin_{\tau \in \mathcal T} \sum^{L-1}_{i=0} \lambda_i,
\end{equation}
where $L$ is the length of the trajectory and $\mathcal T$ is the set of all valid trajectories from the desired start to goal state. We define a valid trajectory as one that is contained within the acceptable region 
\begin{equation}
(x_i, z_i) \in  \mathcal W \setminus \mathcal L_s \ \forall \ (x_i, z_i) \in \tau,
\end{equation}
and whose consecutive states are adjacent grid locations.
Given the discretization of the NFM, we can treat $\mathcal M$ as a graph whose edge weights consist of $\lambda$.
Therefore, we use uniform cost search to obtain $\tau^*$.  Algorithm~\ref{alg:uniform_cost_search} provides the pseudocode of the path planning algorithm.

\begin{algorithm}[t]
\SetAlgoLined
\LinesNumbered
\DontPrintSemicolon
\KwIn{$\bar x_s, \bar z_s, \bar x_g, \bar z_g, \mathcal M$}
\KwOut{$\tau^*$}
\SetKwProg{Fn}{Func}{:}{}
\SetKwFunction{DiscretizeManifold}{DiscretizeManifold}
\SetKwFunction{UniformCostSearch}{UCS}
{
\Fn{\UniformCostSearch{$\bar x_s, \bar z_s, \bar x_g, \bar z_g, \mathcal M$}}
{
$\mathcal W \gets$ valid workspace of $\mathcal M$ \;
$\mathcal L_s \gets l_s$ penalty region  \;
$\mathbf h \gets$ initialize min heap priority queue \;
$\mathbf c \gets$ initialize empty list \;
$n_s \gets$ node with location ($\bar x_s, \bar z_s$) and cost 0 \;
$n_g \gets$ node with location ($\bar x_g, \bar z_g$) and cost 0 \;
$\mathbf h$.push($n_s$) \;
\While{\textup{len($\mathbf h) > $ 0}}{
$n_i \gets$ $\mathbf h$.pop() \;

\If{$n_i == n_g$}
{
    $\tau^* \gets$ path from start to goal \;
    break \;
}
$\mathbf c$.append($n_i$) \;

\For{$(\bar x_j, \bar z_j) \in $ \textup{neighbors of $n_i$}}
{
    \If {$(\bar x_j, \bar z_j) \notin \mathcal W \setminus \mathcal L_s$}
    {
        continue \;
    }
    $n_j \gets$ node with location $(\bar x_j, \bar z_j)$ and cost $\lambda_j$ from $\mathcal M$ \;
    \If{$n_j \in \mathbf c$}
    {
        continue \;
    }
    \If{$n_j \in \mathbf h$ \textup{and cost of} $n_j$ \textup{is higher}}
    {
        continue \;
    }
    $\mathbf h$.push($n_j$) \;
}
}
$\tau^* \gets$ perform trajectory smoothing on $\tau^*$ \;
\textbf{return} $\tau^*$ \;

}
}
\caption{Uniform Cost Search}
\label{alg:uniform_cost_search}
\end{algorithm}

\section{Robotic System}
\label{sec:robotic_system}

\begin{figure}[t]
	\includegraphics[width=\columnwidth]{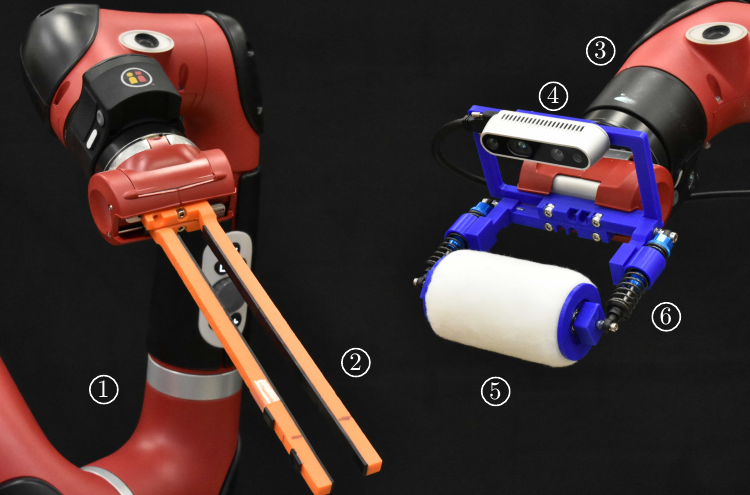}
	\caption{Experimental apparatus: Two robot manipulators, one for folding (1) and the other for creasing (3). An elongated gripper (2) is used to grab the manipulated end of the paper. A roller (5) with compliant springs (6) is used to form the crease. An Intel RealSense D435 camera (4) attached to the creasing arm offers visual feedback during the folding procedure. All gripper attachments were 3D printed.}
	\label{fig:apparatus}
\end{figure}

\begin{figure*}
	\includegraphics[width=\textwidth]{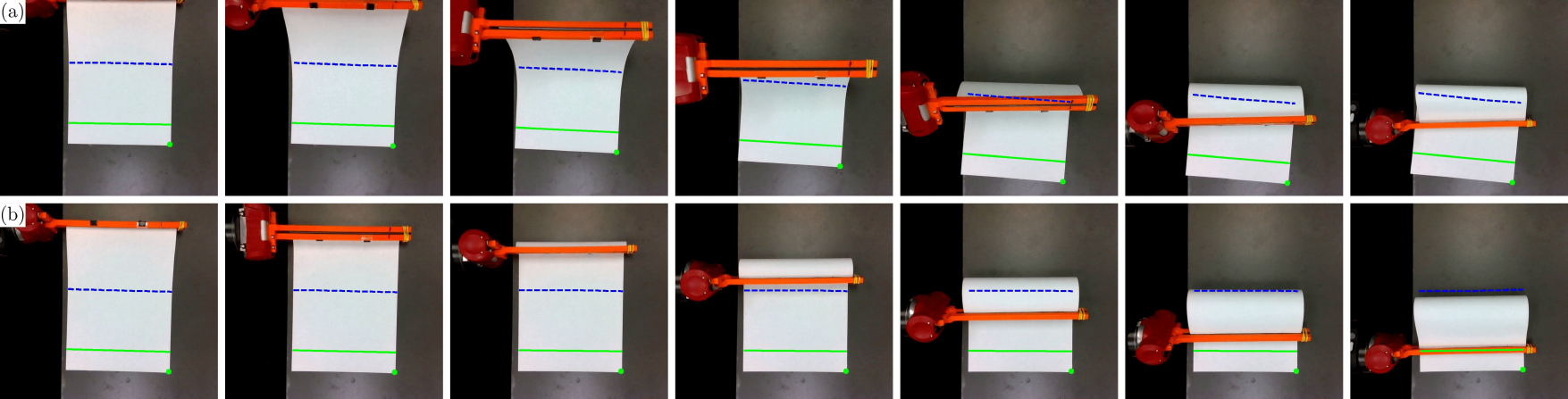}
	\caption{Example of our perception system with a top-down view of the folding procedure. (a) Shows the intuitive baseline results while (b) shows our open-loop algorithm for $L_{gb} = 0.048$ and $C=0.25$\,m. As in Fig.~\ref{fig::Problem_Statement}, the solid green line indicates the desired end effector position while the dashed blue line indicates the crease location. For this case, we observe that the intuitive baseline suffers from considerable sliding while our open-loop algorithm has near-perfect performance.}
	\label{fig:top_down_seq}
\end{figure*}

\subsection{Dual Manipulator Setup}

For our experiments, we use two Rethink Robotics' Sawyer manipulators, as shown in Fig.~\ref{fig:apparatus}. One arm has an elongated gripper designed for folding, while the other arm has a spring-compliant roller for creasing and an Intel Realsense D435 camera for visual feedback. The elongated gripper has rubber attached to the insides of the fingers for tight gripping.

\subsection{Perception System}
\label{subsec:perception_system}

For perception, we take an eye-in-hand approach by attaching an Intel Realsense D435 camera to the roller arm.
We do not use the range output of the camera as it points down along the world $z$-axis and the distance from the camera to the table is known.
To determine the pose of the paper, we use simple color detection to segment the paper and then use Shi-Tomasi corner detection~\cite{shi1994corner} to obtain the position of the bottom edge. Fig.~\ref{fig:top_down_seq} shows an example of the top-down view as well as the poses detected by the vision system. 

\subsection{Model-Predictive Control via Visual Feedback}
\label{sec:feedback-control}

Although we minimize $\lambda$ with our proposed framework, sliding could still occur due to a substrate's low friction surface and/or jittering of the robot's end-effector.
Notice that the optimal trajectory $\tau^*$ generated as described in Sec.~\ref{subsec:path_planning} assumes that the origin $\mathbf o$ of our coordinate system, shown in Fig.~\ref{fig:schematic}(a), is fixed.
We can define the origin as  $\mathbf o = \mathbf q_0 - l \hat{\mathbf x}$, where $l$ is the total length of the paper.
Any amount of sliding indicates that $\mathbf q_0$ is moving along the $x$-axis and, therefore, the origin $\mathbf o$ also moves an identical amount.
When this occurs, our position within the manifold during traversal deviates from the optimal trajectory. Furthermore, without adaptive replanning, the amount of sliding $\Delta x$ will directly result in $\Delta x$ amount of error when creasing.
To circumvent this, we introduce a model-predictive control approach that mitigates the effects of sliding through trajectory corrections via visual feedback.

\begin{figure}[t]
    \centering
    \includegraphics[width=\columnwidth]{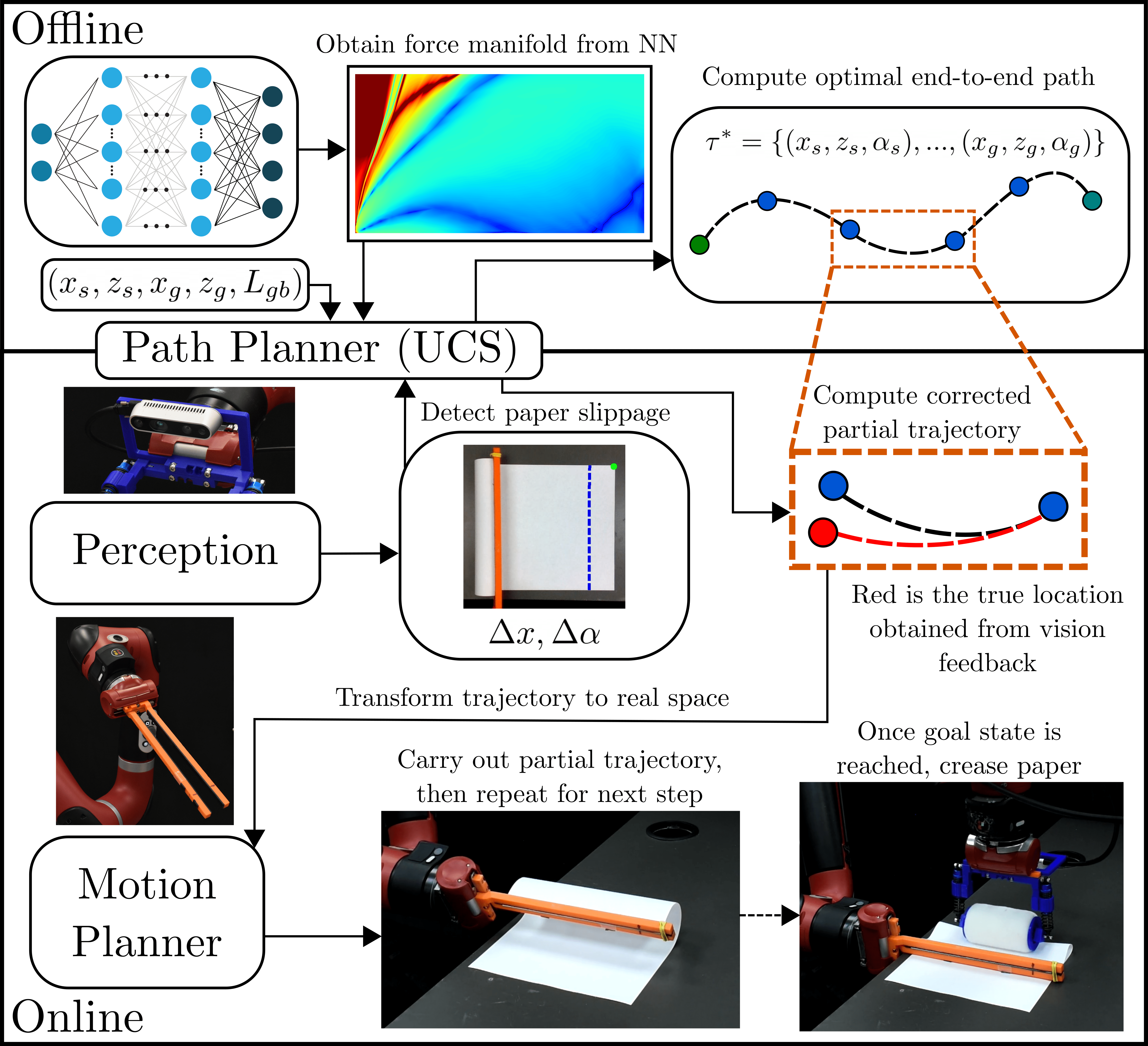}
	\caption{Overview of our robotic paper-folding pipeline. The top row shows offline components while the bottom row shows online ones. On the offline side, we use our trained neural network to generate the necessary force manifold for planning. Then, given an input tuple $(x_s, z_s, x_g, z_g, L_{gb})$, we generate an end-to-end trajectory using uniform cost search. This end-to-end trajectory is then split up into partial trajectories that are carried out by the robot. At the conclusion of each partial trajectory, we measure paper sliding and replan the next partial trajectory to rectify the error.}
	\label{fig:pipeline}
\end{figure}

We acquire visual feedback at $N$ evenly spaced intervals along the trajectory $\tau^*$, as shown in Fig.~\ref{fig:pipeline}. 
To do so, we first partition $\tau^*$ into $N$ partial trajectories.
Aside from the first partial trajectory $\tau^*_0$, we extract the start and goal states of the other $1 \leq i \leq N$ partial trajectories resulting in a sequence of $N$ evenly spaced out states $\mathcal S = \{ (x_1, z_1, \alpha_1),\dots, (x_N, z_N, \alpha_N) \}$ when accounting for overlaps.
After carrying out $\tau^*_0$, we detect the amount of sliding $\Delta x$ and incorporate this error by updating the start state and non-dimensionalizing as 
\begin{equation}
\bar x_i^c = \frac{x_i - \Delta x}{L_{gb}}.
\end{equation}
We then replan a partial trajectory $\tau^*_i$ from the updated start state $(x^c_i, z_i)$ to the next state $(x_{i+1}, z_{i+1})$ in the sequence and carry out this updated trajectory. 
This is repeated until reaching the goal state.
By properly accounting for sliding, we ensure that the traversal through the NFM is as accurate as possible.
Note that this scheme allows us to obtain corrected partial trajectories in near real time once $N$ becomes sufficiently large, as each partial trajectory's goal state approaches its start state, allowing for uniform cost search to conclude rapidly. We refer the reader to our supplementary videos (Footnote~1), which showcase the speed of the feedback loop.

The sliding $\Delta x$ is not the only error we must rectify. Recall that we assume an optimal grasp orientation $\alpha$ for each position within the manifold. When the origin of our NFM moves, the true position does not match the intended position, resulting also in an angular error
\begin{equation}
\begin{aligned}
\alpha^c_i &= \mathcal F_\textrm{NN}(\bar x_i^c, \bar z_i), \\
\Delta \alpha &= \alpha_i - \alpha^c_i.
\end{aligned}
\end{equation}
Simply applying a $-\Delta \alpha$ update to the first point in a partial trajectory results in a large rotational jump that only exacerbates the sliding issue. 
Furthermore, so long as sliding is not extremely large, the incorrect $\alpha$ at the current position within the manifold is still fairly optimal.
Therefore, the $\Delta \alpha$ error is incorporated into the trajectory gradually:
\begin{equation}
\begin{aligned}
\tau^*_i &= \textrm{UCS}(\bar x^c_i, \bar z_i, \bar x_{i+1}, \bar z_{i+1}, \mathcal M), \\
\boldsymbol \alpha_i &= \mathcal F_{\textrm{NN}}(\tau^*_i), \\
\boldsymbol \alpha^c_i &= \boldsymbol \alpha_i + \Delta \alpha[1, (L-1)/L, ..., 1/L, 0]^T,
\end{aligned}
\end{equation}
where UCS denotes uniform cost search and $L$ is the length of trajectory $\tau^*_i$. 
This gradual correction ensures that we minimize sliding while maintaining smoothness of the trajectory.
Algorithm~\ref{alg:pseudo_code} provides the pseudocode for our full closed-loop method.

\begin{algorithm}[t]
\SetAlgoLined
\LinesNumbered
\DontPrintSemicolon
\KwIn{$(x_s, z_s), (x_g, z_g), L_{gb}, \delta, N, \mathcal F_{\textrm{NN}}$}
\SetKwFunction{DiscretizeManifold}{DiscretizeManifold}
\SetKwFunction{UniformCostSearch}{UCS}
\SetKwFunction{SplitTrajectory}{SplitTrajectory}
{
$\mathcal M \gets $\DiscretizeManifold($\mathcal F_{\textrm{NN}}, \delta$) \;
$\bar x_s, \bar z_s, \bar x_g, \bar z_g \gets$ non-dimensionalize with $L_{gb}$ \;
$\bar \tau^* \gets$ \UniformCostSearch($\bar x_s, \bar z_s, \bar x_g, \bar z_g, \mathcal M$) \;
update $\bar \tau^* $ with $\alpha_s$ using $\mathcal F_{\textrm{NN}}$ \;
$\tau^* \gets$ convert $\bar \tau^*$ to real space with $L_{gb}$ \;
$\tau^*_0, ..., \tau^*_{N-1} \gets$ \SplitTrajectory($\tau^*, N$) \;
$\mathcal S \gets$ extract start and goal states\;
carry out $\tau^*_0$ on robot \;
\For{$(x_i, z_i, \alpha_i)$ \textup{and}  $(x_{i+1}, z_{i+1}, \alpha_{i+1}) \in \mathcal S$}
{
    $\Delta x \gets$ detect sliding of paper \;
    $x_i^c \gets x_i - \Delta x$ \;
    $\bar x_i^c, \bar z_i, \bar x_{i+1}, \bar z_{i+1} \gets$ non-dimensionalize with $L_{gb}$ \;
    $\alpha^c_i \gets \mathcal F_{\textrm{NN}}(\bar x_i^c, \bar z_i) $ \;
    $\Delta \alpha \gets \alpha_i - \alpha^c_i$ \;
    $\bar \tau^*_i \gets $ \UniformCostSearch($\bar x_i^c, \bar z_i, \bar x_{i+1}, \bar z_{i+1}, \mathcal M$) \;
    $L \gets$ len($\bar \tau^*_i$) \;
    $\boldsymbol \alpha_i \gets$ obtain $\alpha$s of $\bar \tau^*_i$ using $\mathcal F_\textrm{NN}$ \;
    $\boldsymbol \alpha^c_i \gets \boldsymbol \alpha_i + \Delta \alpha [1, (L-1)/L, ..., 1/L, 0]^T$ \;
    append $\bar \tau^*_i$ with $\boldsymbol \alpha^c_i$ \;
    $\tau^*_i \gets$ convert $\bar \tau^*$ to real space with $L_{gb}$ \;
    carry out $\tau^*_i$ on robot \;
}
crease paper with roller\;
}
\caption{Closed-loop Control Pseudocode}
\label{alg:pseudo_code}
\end{algorithm}

\section{Experiments and Analysis}
\label{sec:experiments_and_analysis}

\subsection{Measuring the Material Property of Paper}

To use our framework, we must develop a way to accurately measure the parameter $L_{gb}$.
Recall that $L_{gb}$ is composed of the bending stiffness $k_b = Eh^3/12$ and density $\rho$.
Therefore, we need only measure this single quantity to describe the paper's material properties.
We next propose a simple way to measure the parameter.

\begin{figure}
	\includegraphics[width=\columnwidth]{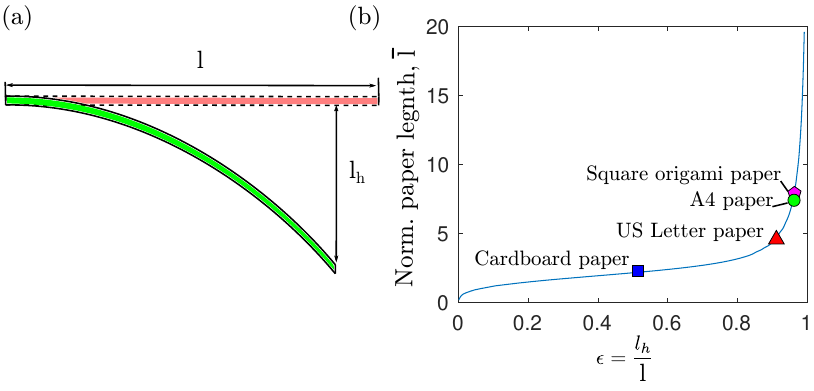}
	\caption{(a) Schematic of a hanging plate. The manipulation edge is fixed horizontally. (b) Relationship between the ratio $\epsilon = l_h/l$ and normalized total length of the paper $\bar l = l/L_{gb}$. }
	\label{fig:measurement}
\end{figure}

As shown in Fig.~\ref{fig:measurement}(a), when one end of the paper is fixed, it will deform due to the coupling of bending and gravitational energy. 
As $L_{gb}$ encapsulates the influence of bending and gravity on the paper, we have the following mapping:
\begin{equation}
    \mathcal{L} (\epsilon) = \bar l = \frac{l}{L_{gb}}, \ \ \epsilon = \frac{l_h}{l},  \\
\end{equation}
where $l_h$ is the vertical distance from the free end to the fixed end and $l$ is the total length of the paper. 
We can obtain the mapping $\mathcal L(\epsilon)$ using numerical simulations (Fig.~\ref{fig:measurement}(b)).
With this mapping known, simple algebra can be performed to obtain $L_{gb}$.
First, we measure the ratio $\epsilon = l_h/l$ for a particular paper to obtain its corresponding normalized total length $\bar l$. 
Then, the value of $L_{gb}$ can be calculated simply by $L_{gb} = l / \bar l$. 
Once we obtain $L_{gb}$, we can now use the non-dimensionlized mapping (\ref{eq:mapping_without_unit}) to find the optimal path for manipulating the paper.

\begin{figure}
\centerline{\includegraphics[width=\columnwidth]{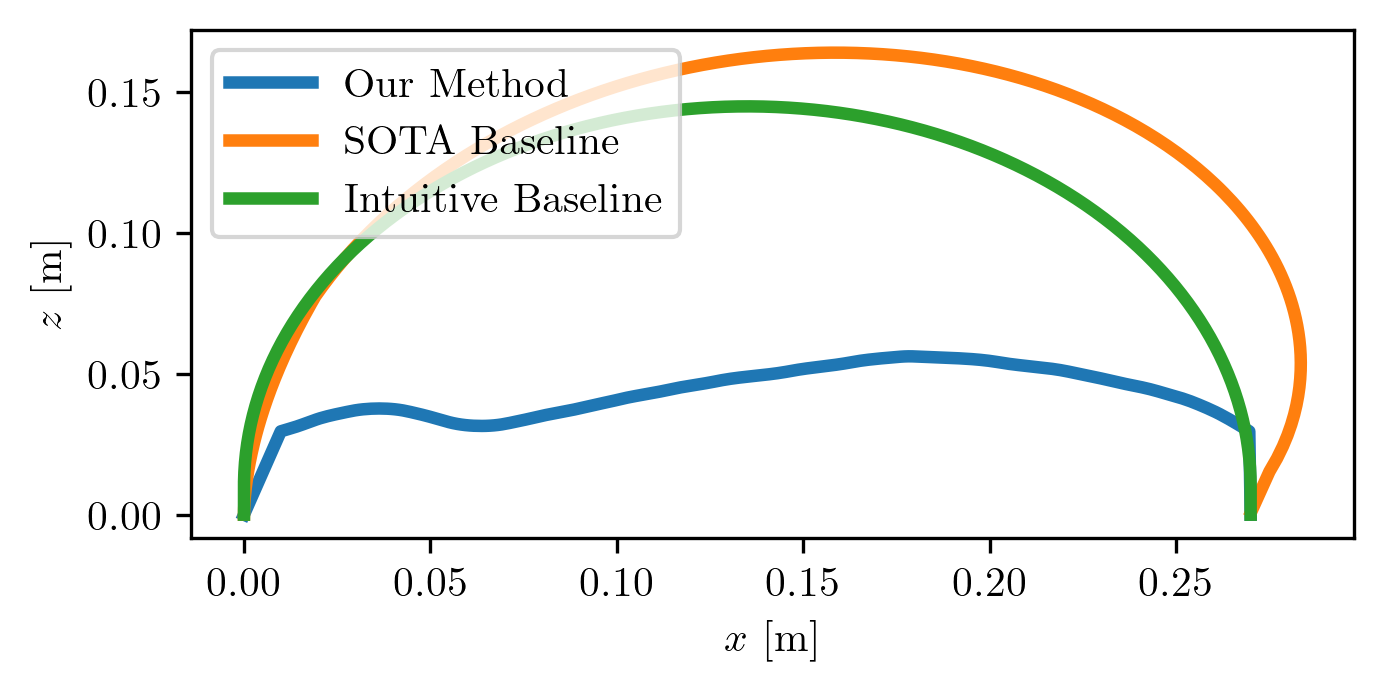}}
\caption{Comparison of trajectories computed by the folding algorithms for US letter paper with $\mathcal C=0.27$\,m.}
\label{fig::baseline_trajs}
\end{figure}

\begin{figure*}
	\includegraphics[width=\textwidth]{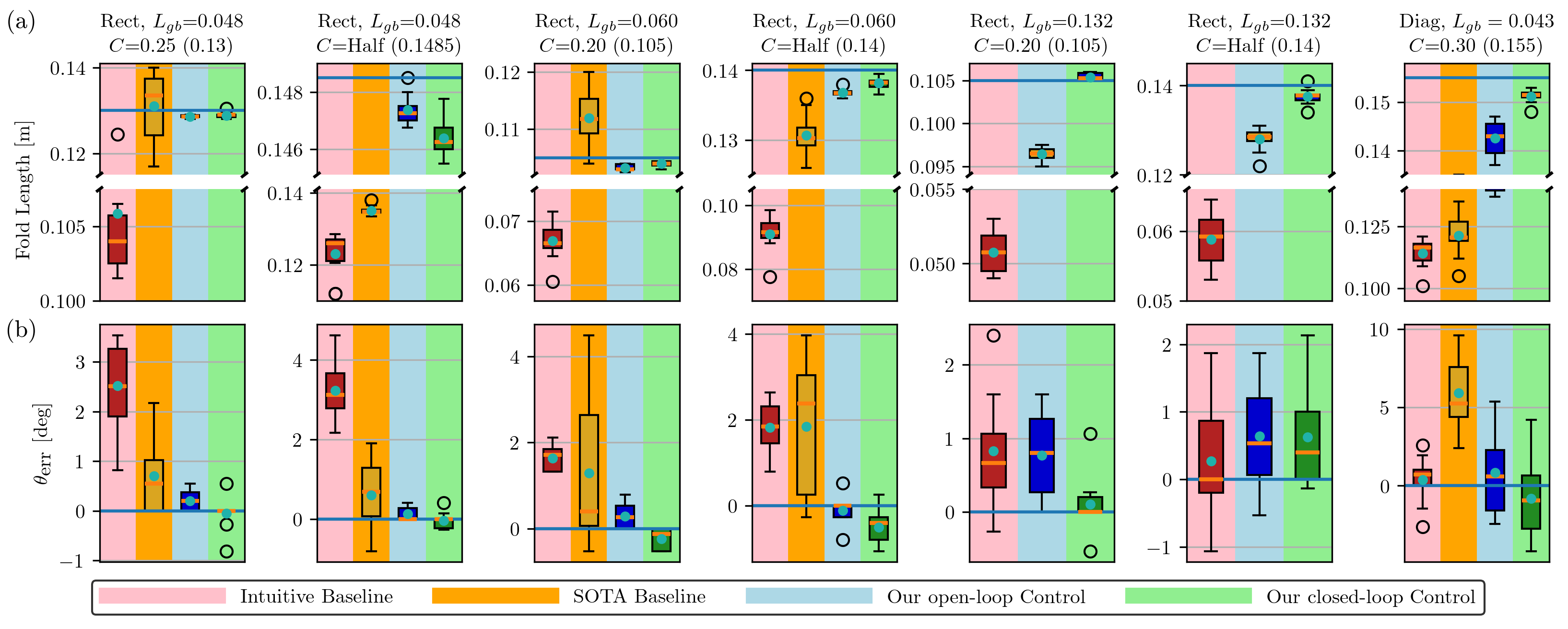}
	\caption{Experimental results for all folding scenarios. Each column indicates a folding scenario while the top row (a) shows the fold length and the bottom row (b) shows the spin error. Boxplot results are shown color coded for the intuitive baseline, the SOTA baseline~\cite{petrik2016folding}, open-loop control, and closed-loop control algorithms. Medians are shown as orange lines, means as turquoise circles, and the desired target value as a light blue horizontal line. Both our open-loop and closed-loop algorithms yield significant improvements over the intuitive baseline and the SOTA baseline, as shown by the broken axis in (a). Our algorithms also exhibit significantly less variance.}
	\label{fig:results}
\end{figure*}

\begin{figure*}
	\includegraphics[width=\textwidth]{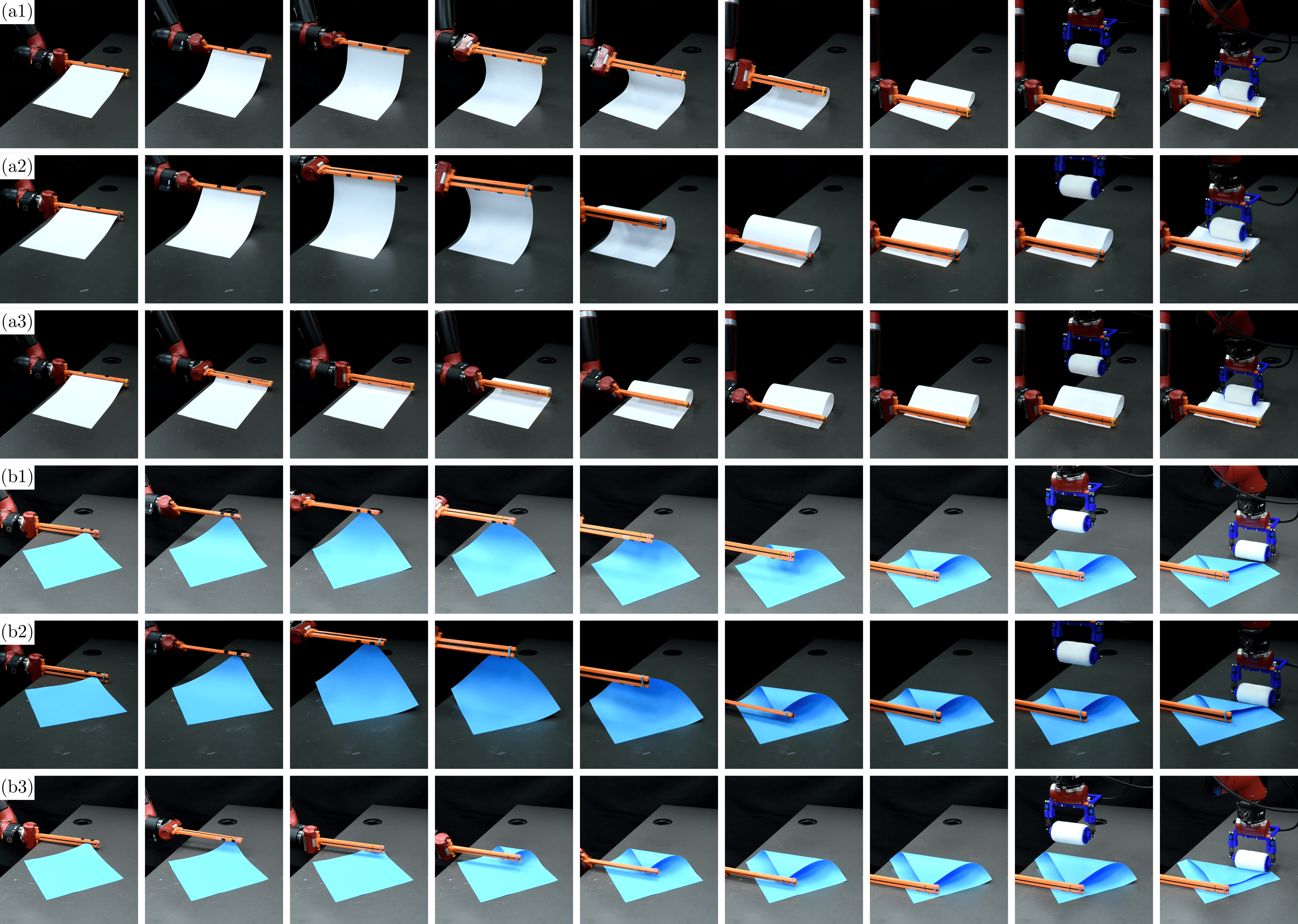}
	\caption{Isometric views of different folding scenarios. (a) $C=\textrm{Half}$ folding for $L_{gb} = 0.048$ paper with the intuitive baseline (a1), the SOTA baseline (a2), and our open-loop algorithm (a3). (b) $C=0.30$\,m diagonal folding for $L_{gb} = 0.043$ with the intuitive baseline (b1), the SOTA baseline (b2), and our closed-loop algorithm (b3).}
	\label{fig:seq1}
\end{figure*}

\subsection{Baseline Algorithms}

To demonstrate the benefits of our folding algorithm, we compared it to both an intuitive and a state-of-the-art baseline. We can think of an intuitive baseline algorithm as one that would work if the opposite end of the paper were fixed to the substrate. Naturally, such a trajectory would be one that grabs the edge of the paper and traces the half perimeter of a circle with radius $R = C / 2$:
\begin{equation}
\begin{aligned}
    \textrm{d}\theta &= \pi / M, \\
    \tau_B &= \{  (R \cos(i \textrm{d}\theta), R \sin(i \textrm{d}\theta), i \textrm{d}\theta), \ \forall i \in [0, M] \},\\
\end{aligned}
\end{equation}
where $M$ is an arbitrary number of points used to sample the trajectory. We chose $M=250$ for all our experiments.

Additionally, we conducted comparisons against the state-of-the-art mechanics-based folding algorithm presented by Petrik et al.~\cite{petrik2016folding, petrik2020static}, to which we refer as the ``SOTA baseline'', which uses a beam model to compute folding trajectories for fabric minimizing sliding. 
However, this baseline considers only the current status of the deformed material when computing subsequent optimal grasp and, consequently, is unable to handle the challenging task of paper folding. 
Examples of the computed trajectories are shown in Fig.~\ref{fig::baseline_trajs}.

\subsection{Experimental Setup}
\label{subsec:experimental_setup}

We tested folding on 4 different types of paper:
\begin{enumerate}
    \item A4 paper, $L_{gb} = 0.048$\,m,
    \item US Letter paper, $L_{gb} = 0.060$\,m,
    \item cardboard paper (US Letter dimensions), $L_{gb} = 0.132$\,m,
    \item square origami paper, $L_{gb} = 0.043$\,m.
\end{enumerate}
For the rectangular papers (1--3), we performed two sets of experiments. The first involved folding the papers to an arbitrary crease location ($C=0.25$\,m for A4 and $C=0.20$\,m for US Letter and cardboard), while the second involves folding the papers in half. For the square origami paper, we chose an arbitrary crease location of $C=0.30$\,m.
This resulted in a total of 7 folding scenarios. For each of the scenarios, we conducted experiments using 4 different algorithms---the intuitive baseline, the SOTA baseline, our open-loop approach, and our closed-loop approach. We completed 10 trials for each of these algorithms, resulting in 280 experiments.

We also validated our model's non-dependence on the paper's width $w$ (Sec.~\ref{sec:reduced_order_model}) through additional experiments involving narrow strip folding. 
We created strips of width $w=2.5$\,cm for both A4 and cardboard and performed 10 trials for each algorithm, resulting in 80 additional experiments for a total of 360 experiments in our extensive case study.

\begin{table*}
\centering
\caption{Offline trajectory computation times for papers and crease types [s]}
\begin{tabular}{lccccccc}
\toprule
\textbf{Algorithm} & A4 $C=0.25$\,m &  A4 $C=$ Half & US $C=0.20$\,m &  US $C=$ Half & CA $C=0.20$\,m &  CA $C=$ Half & OR $C=0.30$\,m\\
\midrule
\textit{Petrik et al.\cite{petrik2016folding, petrik2020static}}   & $59.46$ & $51.15$ & $68.14$ & $47.30$ & $80.07$ & $77.28$ & $43.20$ \\
\textit{Our Method}  & $3.28$ & $4.13$  & $1.80$ & $2.28$ & $1.27$ & $4.19$ & $11.73$ \\
\bottomrule
\end{tabular}
\label{tab:offline_traj_times}
\end{table*}

\begin{figure*}[t]
	\includegraphics[width=\textwidth]{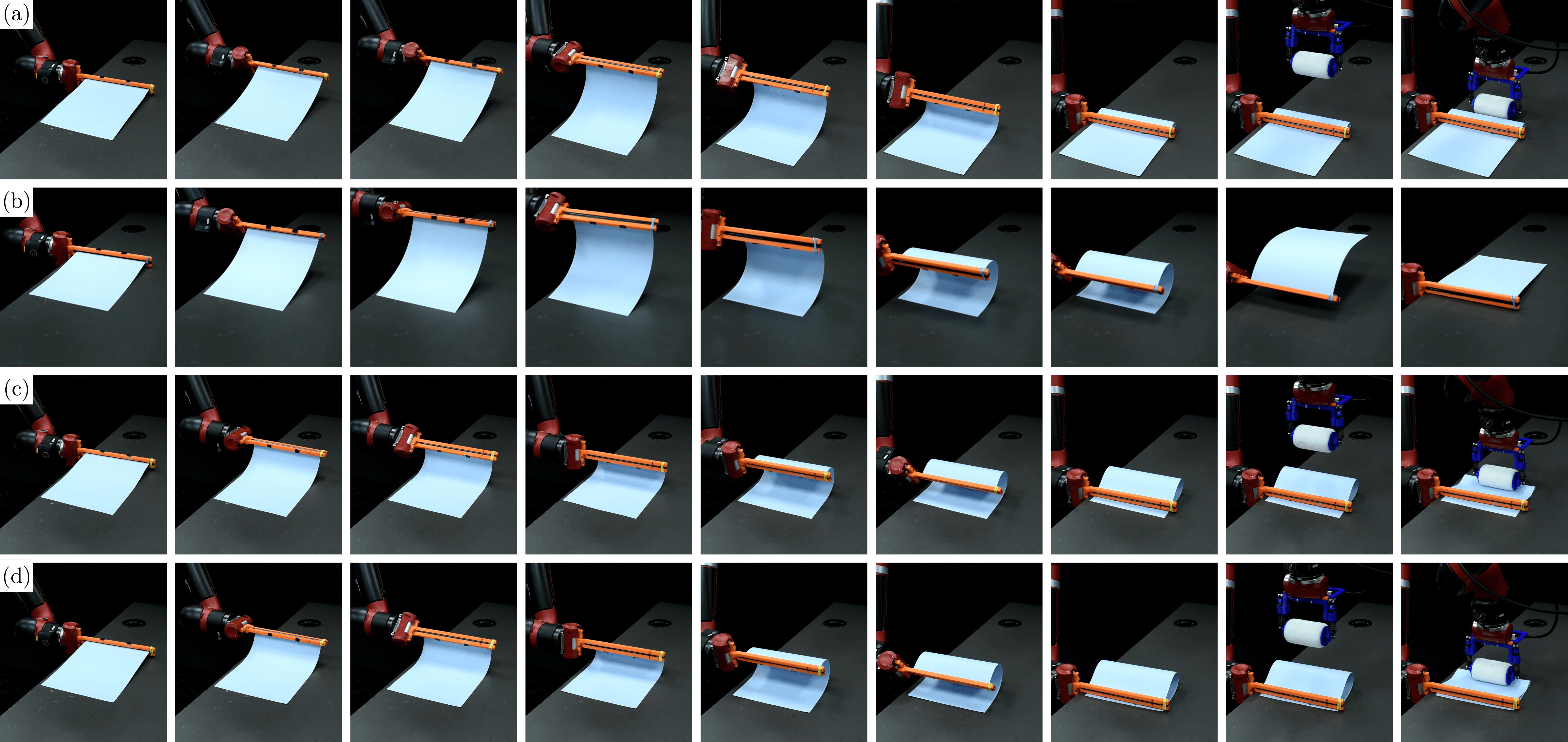}
	\caption{Isometric views for folding $C=\textrm{Half}$ with the stiffest paper ($L_{gb}=0.132$): (a) shows the intuitive baseline, which fails drastically as the stiffness of the paper causes excessive sliding during the folding process, (b) shows the SOTA baseline, which is unable to fold cardboard at all and experiences a high energy snap caused by the large induced deformations, (c) shows our open-loop algorithm, demonstrating significant improvements over both baselines with minimal sliding, and (d) shows our closed-loop algorithm, which improves upon our open-loop results and achieves near perfect folding.}
	\label{fig:seq2}
\end{figure*}

\begin{table*}[t]
\renewcommand{\arraystretch}{1.1}
\centering
\caption{Evaluation of the Influence of Paper Width}
\resizebox{\textwidth}{!}{%
\begin{tabular}{l c cccc cccc}
\toprule
\multirow{2}{*}[-2pt]{Paper $\left(C/2 \right)$} &  \multirow{2}{*}[-2pt]{$w$ [cm]} & \multicolumn{4}{c}{Fold length $\textrm{mean} \pm \sigma$ [cm]} & \multicolumn{4}{c}{Spin error $\theta_\textrm{err}$ $\textrm{mean} \pm \sigma$ [$\deg$]}\\
\cmidrule(lr){3-6}
\cmidrule(lr){7-10}
&  & INT-base &  SOTA-base & Open-loop & Closed-loop &INT-base &  SOTA-base & Open-loop & Closed-loop \\
\midrule
\multirow{2}{*}{A4 (14.85)} & $2.5$  & $10.89 \pm 0.69$ & $14.66 \pm 0.46$ & $14.88 \pm 0.07$& $14.83 \pm 0.04$ & $0.36 \pm 1.65$ & $0.02 \pm 0.65$ & $0.48 \pm 0.47$& $-0.23 \pm 0.48$  \\
  & $21$  & $12.31 \pm 0.63$ & $13.51 \pm 0.13$ & $14.74 \pm 0.05$ & $14.64 \pm 0.07$   & $6.56 \pm 1.29$ & $1.68 \pm 0.97$ & $0.27 \pm 0.44$ & $0.35 \pm 0.38$ \\
\multirow{2}{*}{CB (14.0)} & $2.5$  & $7.00 \pm 0.45$ & N/A & $13.31 \pm 0.14$& $13.81 \pm 0.08$  & $0.07 \pm 0.75 $ & N/A & $-0.70 \pm 0.31$&  $-0.35 \pm 0.82$ \\
 & $21$  & $5.88 \pm 0.38$ & N/A & $12.8 \pm 0.27$ &  $13.75 \pm 0.19$ & $1.25 \pm 1.43$ & N/A & $1.38 \pm 1.06$ &  $1.45 \pm 1.48$ \\
 \bottomrule
\end{tabular}}
\label{tab:eval_width}
\end{table*}

\subsection{Metrics}

The metrics used for the experiments were the average fold length and the spin error. The average fold length was calculated by simply taking the average of the left and right side lengths up until the crease. 
The spin error was calculated as the angle $\theta_\textrm{err}$ that results in the difference between the left and right side lengths.
For square papers, the fold length was defined as the perpendicular length from the tip to the crease and the spin error was the angular deviation from this line to the true diagonal.

\subsection{Parameters}

The neural force manifold $\mathcal M$ was discretized using $\bar \delta=0.0548$ as we found this discretization to be a good compromise between accuracy and computational speed.
All rectangular papers used a penalty region $\mathcal L_s$ defined by $\bar l_s < 0.958$ while the square paper used one defined by $\bar l_s < 1.137$. 
This discrepancy is due to the fact that the diagonal paper has a smaller yield strength compared to the rectangular paper; i.e., to prevent extremely high curvatures, a larger suspended length $\bar l_s$ range must be avoided.

For closed-loop control, we chose to split all trajectories into $N=5$ intervals regardless of trajectory length.
Furthermore, we used a slick (i.e., low friction) table to demonstrate the robustness of our method.
Note that smaller friction coefficients result in a significantly harder manipulation problem due to the lower threshold for sliding.
This is made evident by the excessive sliding of the baseline algorithms shown later.
Using an empirical method, we conducted measurements to determine the static coefficient of friction between the papers and substrate, yielding an approximate value of $\mu_s = 0.12$. 

\subsection{Results and Analysis}

In Table~\ref{tab:offline_traj_times} we report the offline trajectory computation times for all experiments using a single Intel i9-9900KF CPU, demonstrating on average a $15\times$ speed improvement over the SOTA baseline.
In Fig.~\ref{fig:results}, all experimental results for non-narrow strips are reported as box plots where we show the achieved fold lengths and spin errors. From the achieved fold lengths, we see significant improvement over the two baselines for all folding scenarios.
As expected, the SOTA baseline demonstrates better performance compared to the intuitive method with the exception of cardboard paper where the former fails to fold it at all.
Due to the large gap in performance, broken axes are used to properly display the variance of the recorded data. 
Note that not only do our algorithms achieve significantly better performance on average, the variance of our approach is also much lower as shown by the decreased $y$-axis resolution after the axis break.
We attribute the high variances of the baseline algorithms to the increased influence of friction, which can often cause chaotic, unpredictable results.
In other words, truly deterministic folding can only be achieved when sliding is nonexistent.

For the vast majority of cases, incorporating visual feedback yields a clear improvement over the open-loop algorithm.
Intuitively, we observe a trend where the performance gap between our open-loop and closed-loop algorithms grows as the material stiffness increases for rectangular folding.
For softer materials ($L_{gb} = 0.048$), the open-loop algorithm has near perfect performance as shown when folding a paper in half in Fig.~\ref{fig:seq1}(a3). By comparison, Fig.~\ref{fig:seq1}(a1)--(a2) shows the intuitive and SOTA baselines failing with significant sliding.

The sliding problem is only exacerbated by increasing the stiffness of the material ($L_{gb}=0.132$).
Fig.~\ref{fig:seq2}(a) shows the intuitive baseline algorithm failing to fold the cardboard paper in half by a margin almost as long as the paper itself, while Fig.~\ref{fig:seq2}(b) shows how the SOTA baseline method experiences complete failure due to a high energy snapping caused by excessive deformation.
By comparison, our open-loop algorithm is capable of folding the cardboard with significantly better results albeit with some sliding (Fig.~\ref{fig:seq2}(c)). 
As the material stiffness increases, the benefits of visual feedback are more clearly seen as we are able to achieve near perfect folding for cardboard (Fig.~\ref{fig:seq2}(d)).
All of our findings for rectangular folding also match the results of our diagonal folding experiment shown in Fig.~\ref{fig:seq1}(b1)--(b3), where the closed-loop approach once again achieves minimal sliding when compared to the baselines.
Overall, the matching findings across all our experiments demonstrate the robustness of our formulation against material and geometric factors.

We observed one oddity for the folding scenario of $L_{gb}=0.048$ and $C=\textrm{Half}$, in which the open-loop algorithm outperformed our closed-loop variant, but the decrease in performance is on average only 1\,mm, which is attributable to repetitive discretization error caused by $N=5$ replanning.
In fact, as we use a discretization of $\delta = 2$\,mm for the manifold, compounding rounding errors can easily cause 1--2\,mm errors.
With this in mind, our closed-loop method achieves an average fold length performance within a 1-2\,mm tolerance across all experiments.

In terms of spin error, we found that softer materials had the greatest error.
As the surface of the table is not perfectly flat, any amount of sliding will directly result in uneven spin, as shown in Fig.~\ref{fig:seq1}(a).
As the material stiffness increases, the spin errors became more uniform across the methods as the influence of friction is not enough to deform the paper. 
Nevertheless, we can see that our open and closed-loop algorithms resulted in less sliding than the baseline on average.

\subsection{Effects of Paper Width on Folding Performance}

Previously, we mathematically deduced that the paper's width $w$ should have no influence on our folding scheme (Sec.~\ref{sec:reduced_order_model}).
We now validate this claim through experiments.
Table~\ref{tab:eval_width} reports comparisons of fold length and spin error between narrow strips and wide sheets of paper error for half folding A4 paper and cardboard.
As expected, both our open-loop and closed-loop methods have near identical performance regardless of paper width. 
Aside from a slight exception for open-loop narrow cardboard folding that actually benefits from a $5$\,mm improvement, differences are almost imperceptible to the human eye.
Interestingly, both baseline methods experience significant changes in fold length ($> \pm 1$\,cm) when width is changed.
The prevention of such nondeterministic behavior is another benefit of our method. 

\section{Additional Discussion}
\label{sec:additional_discussion}

\subsection{Performing Multiple Folds on the Same Paper}

\begin{figure*}[t]
	\includegraphics[width=\textwidth]{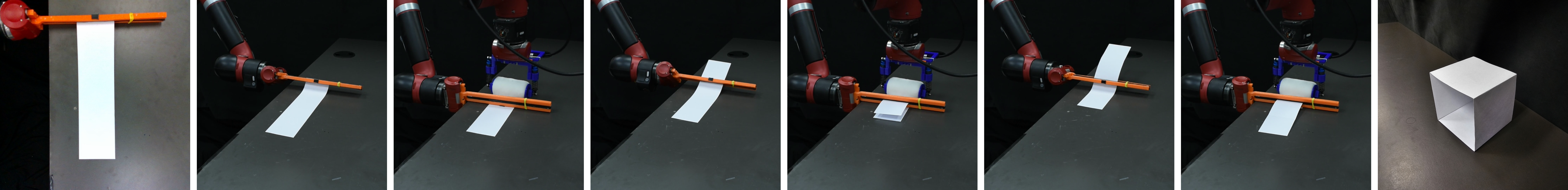}
	\caption{Folding a simple origami cube with open-loop control. The left image shows the starting strip of paper (A4), the middle images show three folding steps to create each corner, and the right image shows the resulting cube. Note that regrasps were performed manually.}
	\label{fig:cube_origami}
\end{figure*}

Our optimal folding strategy can also be used to fold a single piece of paper multiple times so long as our assumptions regarding material homogeneity and symmetrical centerline hold. Given our method's exceptional accuracy, a precise equilateral origami cube can be created using solely open-loop control, and excellent results are shown in Fig.~\ref{fig:cube_origami}. 
Future work pertaining to performing arbitrary folds is discussed in Sec.~\ref{sec::conclusion}.

\subsection{Importance of Single-Manipulator Folding}

\begin{figure*}[t]
	\includegraphics[width=\textwidth]{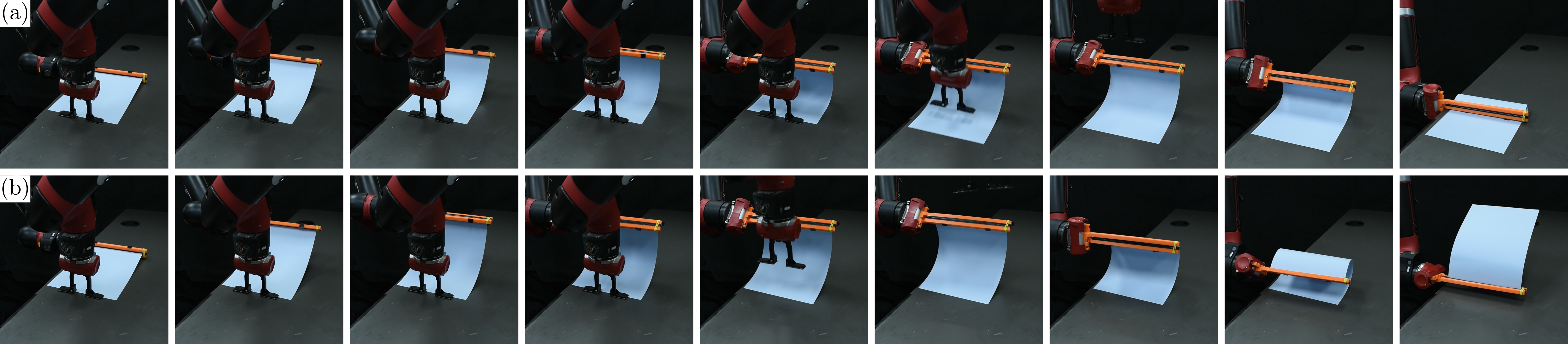}
	\caption{Isometric views for folding $C=\textrm{Half}$ with the stiffest paper ($L_{gb}=0.132$) using an auxiliary manipulator: (a) shows the intuitive baseline while (b) shows the SOTA baseline. Note the similarity in results to Fig.~\ref{fig:seq2} despite the use of an auxiliary manipulator to prevent sliding.}
	\label{fig:fling_out}
\end{figure*}

To emphasize the significance of single-manipulator folding, we next present results for half-folding cardboard using both baselines with the incorporation of an auxiliary manipulator used to prevent the opposing edge from sliding (Fig.~\ref{fig:fling_out}).
The auxiliary manipulator ``holds down'' the edge during the folding process and then moves out of the way at the last possible point in the folding trajectory in order to avoid collision.
Despite the additional hardware, we can see that the results for both baselines are near identical to the original experiments shown in Fig.~\ref{fig:seq2}.

For the intuitive baseline, significant sliding occurs as soon as the auxiliary manipulator lifts away due to the paper's high energy state, ultimately yielding an end result that is nowhere near the desired half-fold.
This sliding is less noticeable for the SOTA baseline, but the sheet still suffers from buckling.
With this demonstration, we conclude that for stiff materials, naive folding strategies can fail drastically despite the use of auxiliary manipulators to explicitly prevent sliding.
In fact, the use of auxiliary manipulators is a rather expensive approach, both in terms of requiring additional hardware as well as introducing more overall motions into the pipeline, which our folding strategy completely avoids with its inherent sliding prevention.

\section{Conclusion}
\label{sec::conclusion}

We have introduced a novel sim2real robot control strategy capable of robustly folding sheets of paper of varying materials and geometries along symmetrical centerlines with only a single manipulator.
Our framework incorporates a combination of techniques spanning several disciplines, including physical simulation, machine learning, scaling analysis, and path planning.
The effectiveness of the framework was demonstrated through extensive real world experiments against both natural and state-of-the-art paper folding strategies.
Furthermore, an efficient, near real-time visual feedback algorithm was implemented that further minimizes folding error.
Our closed-loop model-predictive control algorithm successfully accomplished challenging scenarios such as folding stiff cardboard with highly consistent accuracy.

In future work, we hope to tackle the difficult problem of creating arbitrary creases along sheets of paper with non-symmetric centerlines. Such non-symmetric paper sheets can no longer be represented as a reduced-order 2D elastic rod model, hence requiring a more sophisticated shell-based formulation.
Additionally, precisely folding paper with preexisting creases and folds will be a crucial step to accomplishing elaborate folding tasks, such as robotic origami. 
We believe that our optimal symmetrical folding trajectories can serve as a valuable ``seed'' or initial guess when optimizing for asymmetrical folds.
Moving forward, we anticipate exploring solutions that take advantage of generalized problem formulations with data-driven control schemes, such as reinforcement learning.

\bibliography{arxiv}

\begin{IEEEbiography}[{\includegraphics[width=1in,height=1.25in,clip,keepaspectratio]{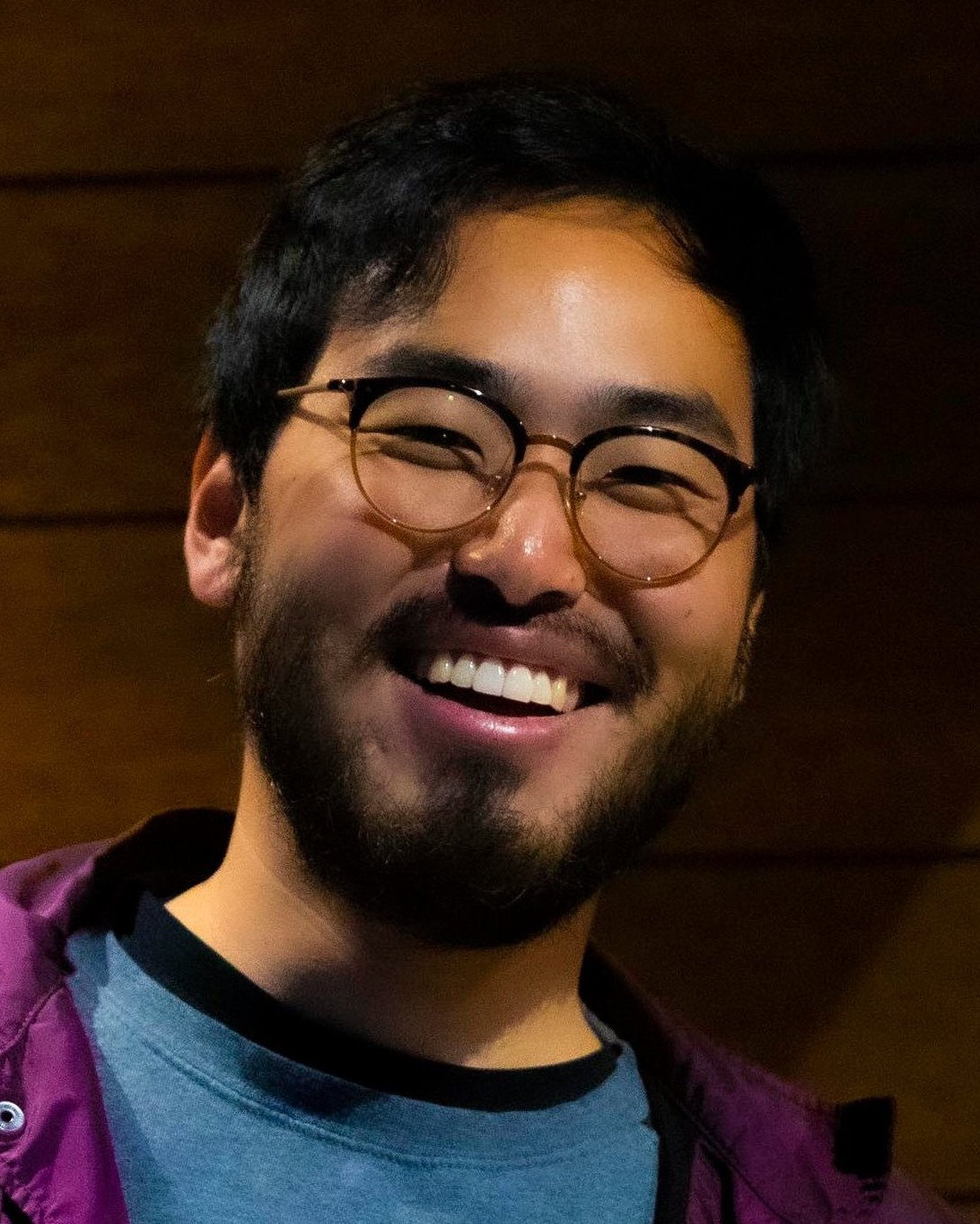}}]{Andrew Choi}
received the B.S. degree in mechanical engineering from the University of California, Davis in 2018 and the Ph.D. degree in computer science from the University of California, Los Angeles in 2023. He is currently a Research Scientist with the Horizon Robotics' General AI Laboratory. His research interests include developing efficient sim2real strategies for robot learning and manipulation. \end{IEEEbiography}

\begin{IEEEbiography}[{\includegraphics[width=1in,height=1.25in,clip,keepaspectratio]{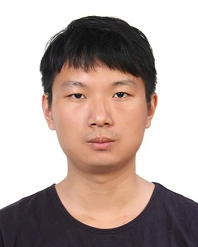}}]{Dezhong Tong} received the B.S. degree in mechanical engineering from Shanghai Jiao Tong University in 2018 and the Ph.D. degree in mechanical engineering from the University of California, Los Angeles in 2023. He is currently a Post-Doctoral Research Fellow with the University of Michigan, Ann Arbor, MI, USA. His research interests include computational mechanics, robotic manipulation, machine learning, and their practical applications in autonomous manufacturing and intelligent systems.\end{IEEEbiography}

\begin{IEEEbiography}[{\includegraphics[width=1in,height=1.25in,clip,keepaspectratio]{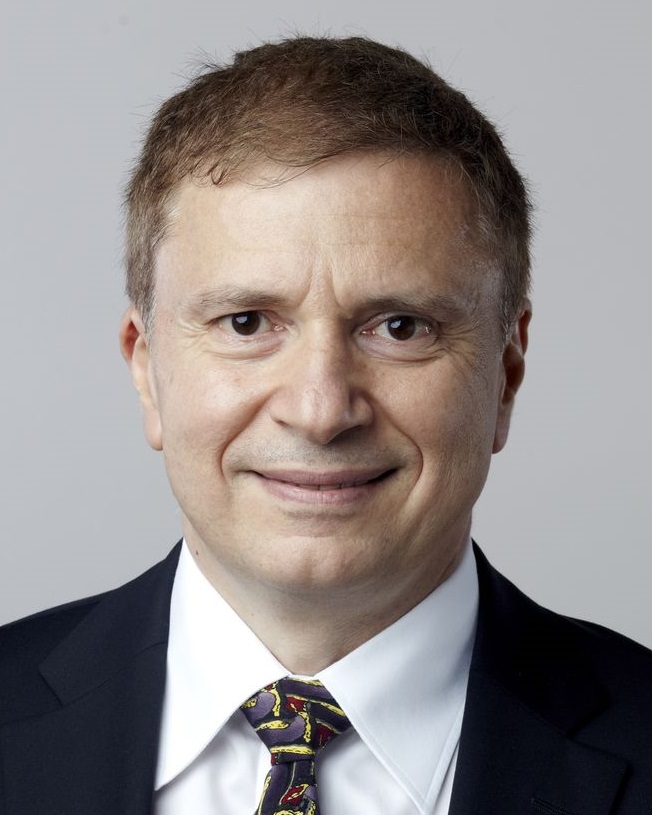}}]{Demetri Terzopoulos}
(Fellow, IEEE) received the Ph.D. degree in artificial intelligence from the Massachusetts Institute of Technology (MIT), Cambridge, MA, USA, in 1984. He is currently a distinguished professor and chancellor’s professor of computer science at the University of California, Los Angeles, Los Angeles, California, where he directs the UCLA Computer Graphics \& Vision Laboratory, and is co-founder and chief scientist of VoxelCloud, Inc. He is or was a Guggenheim fellow, a fellow of the ACM, IETI, Royal Society of Canada, and Royal Society of London, and a member of the European Academy of Sciences, the New York Academy of Sciences, and Sigma Xi. Among his many awards are an Academy Award from the Academy of Motion Picture Arts and Sciences for his pioneering work on physics-based computer animation, and the Computer Pioneer Award, Helmholtz Prize, and inaugural Computer Vision Distinguished Researcher Award from the IEEE for his pioneering and sustained research on deformable models and their applications. Deformable models, a term he coined, is listed in the IEEE Taxonomy.
\end{IEEEbiography}

\vfill\eject

\begin{IEEEbiography}[{\includegraphics[width=1in,height=1.25in,clip,keepaspectratio]{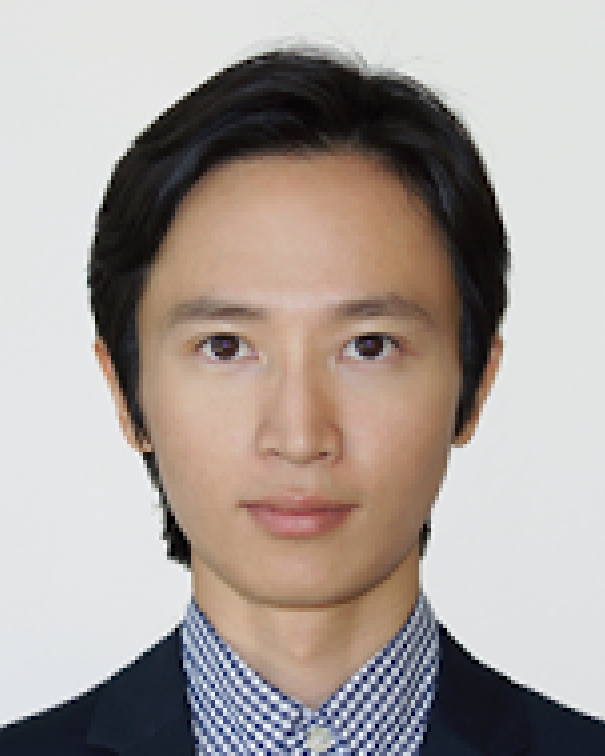}}]{Jungseock Joo}
received the Ph.D. degree in computer science from the University of California, Los Angeles (UCLA) in 2015.
He is currently an Associate Professor of Communication and Statistics with UCLA. His research has addressed a broad range of research questions and practical issues about AI in both technical and social aspects. These topics include Human-AI interaction, face and gestures, collaborative robots, simulations, explainable and fair computer vision and deep learning.\end{IEEEbiography}

\begin{IEEEbiography}[{\includegraphics[width=1in,height=1.25in,clip,keepaspectratio]{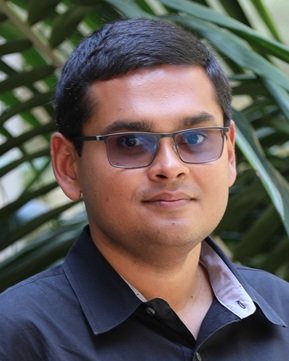}}]{M. Khalid Jawed}
received the Ph.D. degree in mechanical engineering from Massachusetts Institute of Technology in 2016. He is an Associate Professor in the Department of Mechanical and Aerospace Engineering at the University of California, Los Angeles. His current research interests include structural mechanics and robotics.\end{IEEEbiography}

\vfill\eject

\end{document}